\documentclass{article}

     \usepackage[nonatbib, preprint]{neurips_2020}
\usepackage[utf8]{inputenc} % allow utf-8 input
\usepackage[T1]{fontenc}    % use 8-bit T1 fonts
\usepackage{hyperref}       % hyperlinks
\usepackage{url}            % simple URL typesetting
\usepackage{booktabs}       % professional-quality tables
\usepackage{amsfonts}       % blackboard math symbols
\usepackage{nicefrac}       % compact symbols for 1/2, etc.
\usepackage{microtype}      % microtypography

\usepackage{algorithm}
\usepackage{algorithmic}
\newcommand\Tau{\mathcal{T}}
\usepackage{amsmath,amssymb}
\usepackage{graphicx}
\usepackage{booktabs}
\usepackage{dsfont}
\usepackage[super]{nth}

\DeclareMathOperator{\clip}{clip}
\usepackage{etoolbox}
\usepackage{hyperref}

\title{Episodic Self-Imitation Learning with Hindsight}

\author{%
  Tianhong Dai, Hengyan Liu, and Anil Anthony Bharath\\
  Imperial College London, UK\\
  \small\texttt{\{tianhong.dai15, hengyan.liu15, a.bharath\}@imperial.ac.uk} \\
}

\begin{document}

\maketitle

\begin{abstract}
  {Episodic self-imitation learning, a novel self-imitation algorithm with a trajectory selection module and an adaptive loss function, is proposed to speed up reinforcement learning.} Compared~to the {original self-imitation learning algorithm}, which samples good state--action pairs from the experience replay buffer, our agent leverages entire episodes with hindsight to aid self-imitation learning. A selection module is introduced to filter uninformative samples from each episode of the update. {The proposed method} overcomes  the limitations of the standard self-imitation learning algorithm, a transitions-based method which performs poorly in handling continuous control environments with sparse rewards. {From the experiments, episodic self-imitation learning is shown to perform better than baseline on-policy algorithms, achieving  comparable performance to state-of-the-art off-policy algorithms in several simulated robot control tasks. The trajectory selection module is shown to prevent the agent learning undesirable hindsight experiences. With the capability of solving sparse reward problems in continuous control settings, episodic self-imitation learning has the potential to be applied to real-world problems that have continuous action spaces, such as robot guidance and~manipulation. The code is available from here: \href{https://github.com/TianhongDai/esil-hindsight}{https://github.com/TianhongDai/esil-hindsight}.}
\end{abstract}

\section{Introduction}
% sparse rewards
Reinforcement learning (RL) has has been shown to be very effective in training agents within gaming environments~\cite{mnih2015human,silver2016mastering},
%robotic manipulations~\cite{arulkumaran2017deep},
particularly when combined with deep neural networks~\cite{lecun2015deep,silver2016mastering,liu2017survey}. 
In most tasks settings that are solved by RL algorithms, reward shaping is an essential requirement for guiding the learning of the  agent. 
%Reward functions need to be designed carefully with a lot of domain knowledge according to different types of tasks~\cite{ng1999policy}. 
Reward shaping, however, often requires significant quantities of domain knowledge that {are} highly task-specific~\cite{ng1999policy} {and, even with careful design, can lead to undesired policies}. Moreover, for complex robotic manipulation tasks, manually designing  reward shaping functions to guide the learning agent becomes  intractable~\cite{arulkumaran2017deep,florensa2017reverse} if even minor variations to the task are introduced. For such settings, the application of deep reinforcement learning requires algorithms that can learn from unshaped, and usually sparse, reward signals. The complicated dynamics of robot manipulation exacerbate the difficulty posed by sparse rewards, especially for on-policy RL algorithms. For example, achieving goals that require {successfully} executing multiple steps over a long horizon involves high dimensional control {that must also generalise to work across variations in the environment for each step}. These aspects of robot control result in a situation where a naive RL agent so rarely receives a reward at the start of training that it is not able to learn at all. A common solution in the robotics community is to collect a sufficient quantity of expert demonstrations, {then} use imitation learning to train the agent. However, in some scenarios, {demonstrations are expensive to collect and the achievable performance of a trained agent is restricted by their quantity}. One solution is to use the valuable past experiences of the agent to enhance training, and this is particularly useful in sparse reward~environments.

To alleviate the problems associated with having sparse rewards, there are two kinds of approaches: imitation learning and hindsight experience replay (HER). First, the standard approach of imitation learning is to use supervised learning algorithms and minimise a surrogate loss with respect to an oracle.
The most common form is learning from demonstrations~\cite{hester2018deep,gao2018reinforcement}.
%For example, DQfD~\cite{hester2018deep} learns from small amounts of demonstrations. 
Similar techniques are applied to robot manipulation tasks~\cite{rajeswaran2017learning,vevcerik2017leveraging,nair2018overcoming, james2018task}. When the demonstrations are not attainable, self-imitation learning (SIL)~\cite{oh2018self}, which uses past good experiences (episodes in which the goal is achieved), can~be used to enhance exploration or speed up the training of the agent. {Self-imitation learning} works well in discrete control environments, such as Atari Games. Whilst being able to learn policies for continuous control tasks with dense or delayed rewards \cite{oh2018self}, {the present} experiments suggest that SIL struggles when rewards are sparse. Recently, {hindsight experience replay} has been proposed to solve such goal-conditional, sparse reward problems.
The main idea of HER~\cite{andrychowicz2017hindsight} is that during replay, the~selected transitions are sampled from state--action pairs derived from {achieved goals}%MDPI: Please confirm whether the italic format is necessary? If not, please remove it.
 that are substituted for the real goals of the task; this increases the frequency of positive rewards. {Hindsight experience replay} is used with off-policy RL algorithms, such as DQN~\cite{mnih2015human} and DDPG~\cite{lillicrap2015continuous}, for experience replay and has several extensions~\cite{schaul2015prioritized,liu2018competitive}. {The present experiments} show that simply applying HER with SIL does not lead to an agent capable of performing tasks from the Fetch robot environment. In summary, self-imitation learning with on-policy algorithms for tasks that require continuous control, and for which rewards are sparse, remains unsolved.

\begin{figure}[H]
\centering
\includegraphics[width=0.95\textwidth]{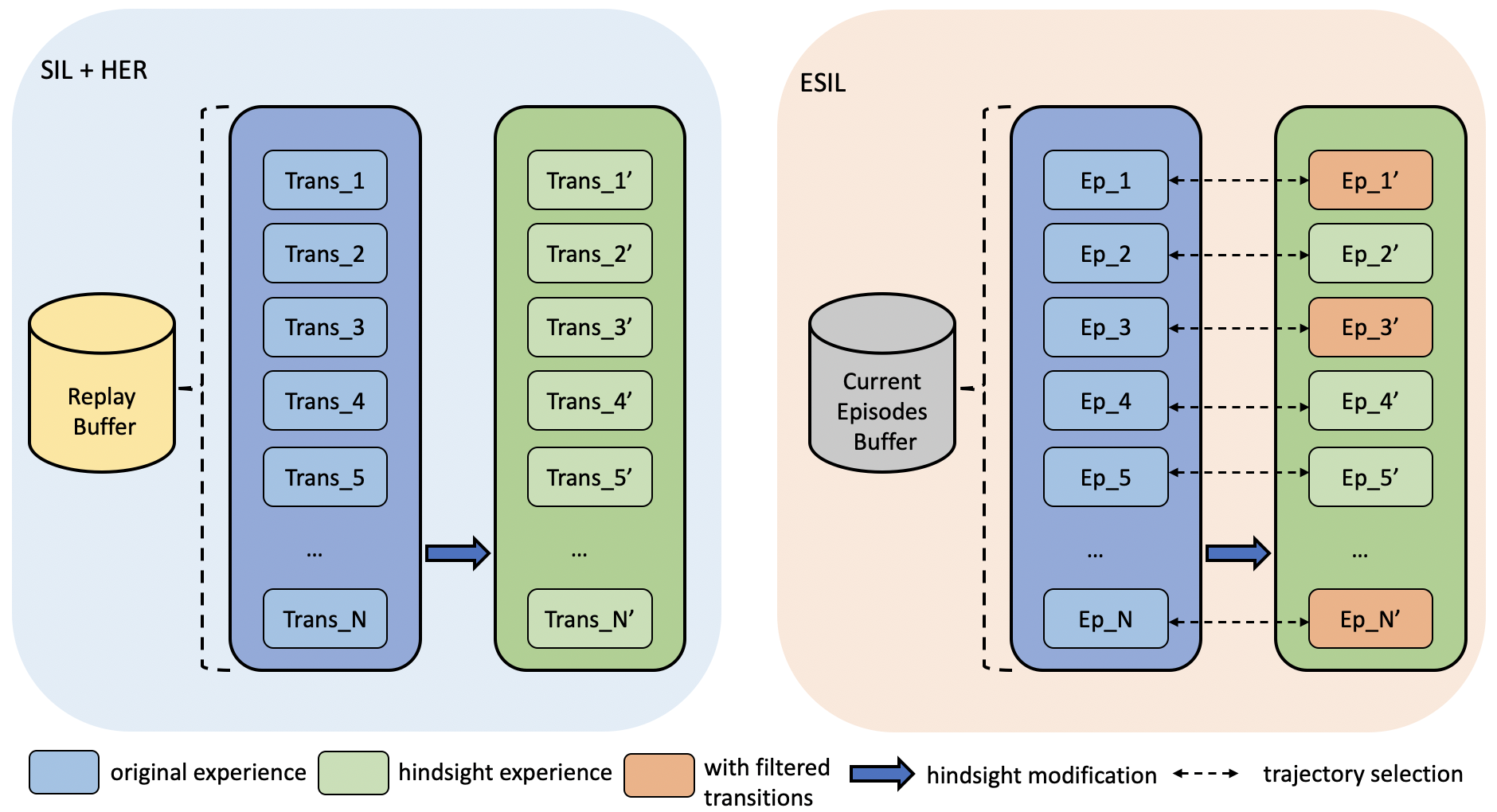}
\caption{Illustration of difference between self-imitation learning (SIL)+hindsight experience replay (HER) and episodic self-imitation learning (ESIL).}
% PPO+SIL+HER samples a batch of transitions from the replay buffer and modify them into hindsight experiences and conduct self-imitation learning directly. PPO+ESIL utilizes entire current collected episodes and convert them into hindsight episodes. Another trajectory selection module is used to remove undesired transitions in each hindsight episodes.
\label{fig:esil}
\end{figure}

In this paper, {episodic self-imitation learning (ESIL) for goal-oriented problems that provide only sparse rewards is proposed and combined with a state-of-the-art on-policy RL algorithm}: proximal policy optimization (PPO). In contrast to standard SIL, which samples past good transitions from the replay buffer for imitation learning, {the proposed} ESIL adopts entire current episodes {(successful or not)}, and modifies them into ``expert'' trajectories based on HER. An extra trajectory selection module is also introduced to relieve the effects of sample correlation~\cite{lee2019sample} in updating the network. {Figure~\ref{fig:esil}} shows the difference between naive SIL+HER and ESIL. {During training by SIL+HER, a batch of transitions is sampled from the replay buffer; these are modified into ``hindsight experiences'' and used directly in self-imitation learning. {In contrast}, ESIL utilises entire current collected episodes and converts them into hindsight episodes. The trajectory selection module removes undesired transitions in the hindsight episodes.} {Using tasks from the Open AI Fetch environment, this paper demonstrates that the proposed ESIL approach is effective in training agents} {which are required to solve continuous control problems}, and shows that it achieves state-of-the-art results on several tasks.

The primary contribution of this paper is a novel episodic self-imitation learning (ESIL) algorithm that can solve continuous control problems {in environments providing only sparse rewards}; in doing so, it also empirically answers an open question posed by \cite{plappert2018multi}. {The proposed ESIL approach also provides} a more efficient way to perform exploration in goal-conditional settings than the standard self-imitation learning algorithm. Finally, this approach achieves, to our knowledge, the best results for four moderately complex robot control tasks in simulation. The paper is organised into the following structure:  Sections~\ref{sec:related_work}  and ~\ref{sec:background} provide an introduction to related work and corresponding background. Section ~\ref{sec:method} describes the methodology of the proposed ESIL approach. Section~\ref{sec:experiment} introduces the settings and results of the experiments. Finally, Section~\ref{sec:conclusion} provides concluding remarks and suggestions for future research. 
%{The code presenting the implementation of ESIL used in this work is available from here: \href{https://github.com/TianhongDai/esil-hindsight}{https://github.com/TianhongDai/esil-hindsight}.}

\section{Related Work}
\label{sec:related_work}
% exploration
Imitation learning (IL) can be divided into two main categories: behavioural cloning and inverse reinforcement learning~\cite{hussein2017imitation}. Behavioural cloning involves the learning of behaviours from demonstrations~\cite{bojarski2016end,xu2017end,torabi2018behavioral}.
Other extensions have an expert in the loop, such as DAgger~\cite{ross2011reduction}, or use an adversarial paradigm for the behavioural cloning method~\cite{ho2016generative,wang2017robust}. The inverse reinforcement learning estimates a reward model from expert trajectories~\cite{ng2000algorithms,abbeel2004apprenticeship,ziebart2008maximum}. Learning from demonstrations is powerful for complex robotic manipulation tasks~\cite{finn2016guided,zhang2018deep,pmlr-v78-finn17a,rajeswaran2017learning,fang2019survey}. ~\cite{ho2016generative} propose generative adversarial imitation learning (GAIL), which employs generative adversarial training to match the distribution of state--action pairs of demonstrations. Compared with behavioural cloning, the GAIL framework shows strong improvements in continuous control tasks. In the work of~\cite{ding2019goal}, goalGAIL is proposed to speed up the training in goal-conditional environments; goalGAIL was also shown to be able to learn from demonstrations without action information. Prior work has used demonstrations to accelerate  learning~\cite{rajeswaran2017learning,vevcerik2017leveraging,nair2018overcoming}.
Demonstrations are often collected by an expert policy or human actions.
In contrast to these approaches, {episodic self-imitation learning (ESIL) does not need demonstrations.}

% self-learning and self-imitation learning
Self-imitation learning (SIL)~\cite{oh2018self} is used for exploiting past experiences for parametric policies. It~has a similar flavor to ~\cite{gangwani2018learning, pmlr-v97-wu19a}, in that the agent learns from imperfect demonstrations. During~training, past good experiences are stored in the replay buffer. When SIL starts, transitions are sampled from the replay buffer according to the advantage values. {In the work of~\cite{tang2020self}, generalised~SIL was proposed as an extension of SIL. It uses an $n$-bound $Q$-learning approach to generalise the original SIL technique, and shows robustness to a wide range of continuous control tasks. Generalised SIL can also be combined with both deterministic and stochastic RL algorithms. \cite{guo2019self} points out that using imitation learning with past good experience could lead to a sub-optimal policy. Instead of imitating past good trajectories, a trajectory-conditioned policy~\cite{guo2019self} is proposed to imitate trajectories in diverse directions, encouraging  exploration in environments where exploration is otherwise difficult.} Unlike SIL, episodic self-imitation learning (ESIL) applies HER to the current episodes to create ``imperfect'' demonstrations for imitation learning; this also requires introducing a trajectory-selection module to reject undesired samples from the hindsight experiences. In the work of ~\cite{lee2019sample}, it was shown that the agent benefits from using whole episodes in updates, rather~than uniformly sampling the sparse or delayed reward environments. {The present} experiments suggest that episodic self-imitation learning {achieves better performance in an agent that must learn to perform continuous control in environments delivering sparse rewards}.

% Hindsight
Recently, the technique known as hindsight learning was developed. Hindsight experience replay (HER)~\cite{andrychowicz2017hindsight} is an algorithm that can overcome the exploration problems in multi-goal environments, delivering sparse rewards. Hindsight policy gradient (HPG)~\cite{rauber2018hindsight} introduces techniques that enable the learning of goal-conditional policies using hindsight experiences.~However,~the~current implementation of HPG has only been evaluated for agents that need to perform discrete actions, and one drawback of hindsight policy gradient estimators is the computational cost because of the goal-oriented sampling. 
An extension of HER, called dynamic hindsight experience replay~(DHER)~\cite{fang2018dher}, was proposed to deal with dynamic goals. \cite{liu2019hindsight} uses the GAIL framework~\cite{ho2016generative} to generate trajectories that are similar to hindsight experiences; it then applies imitation learning, using these trajectories.
Competitive Experience Replay (CER) complements HER by introducing a
competition between two agents for exploration~\cite{liu2018competitive}.
\cite{zhao2018energy} point out that the hindsight trajectories which contain higher energy are more valuable during training, leading to a more efficient learning system.~\cite{fang2019curriculum} proposed curriculum-guided HER, which~ incorporates curriculum learning in the work. During training, the agent focuses on the closest goals in the initial stage, then focuses on the expanding the diversity of goals. This approach accelerates training compared with other baseline methods. Unlike these works, {episodic self-imitation learning (ESIL) combines} episodic hindsight experiences with imitation learning, which aids learning at the start of training. Furthermore, ESIL can be applied to continuous control, making it more suitable for control problems that demand greater precision.

\section{Background}
\label{sec:background}
%\vspace{-6pt}
\subsection{Reinforcement Learning}
{Reinforcement Learning (RL) can be formulated under the framework of a Markov Decision Process (MDP); it is used to learn an optimal policy to solve sequential decision-making problems. In~each time step $t$, the state $s_{t}$ is received by the agent from the environment}. An action $a_{t}$ is sampled by the agent according to its policy $\pi_{\theta}(s_{t}|a_{t})$, parameterised by $\theta$, which---in deep reinforcement learning---represent the weights of an artificial neural network. Then, the state $s_{t+1}$ and reward $r_{t+1}$ are provided by the environment to the agent. The goal is to have the agent learn a policy that maximises the expected return $\mathbb{E}_{\theta}[R(s_t, a_t)]$~\cite{sutton2018reinforcement}
\begin{equation}
\centering
    R(s_t, a_t)=\sum_{l=0}^{T-1}\gamma^{l}r_{t+l},
\end{equation}
where $\gamma$ is the discount factor. {In a robot control setting, the state $s_{t}$ can be the velocity and position of each joint of the robotic arm. The action $a_{t}$ can be the velocities of actuators (control signals)} {and the reward $r_{t}$} {might be calculated based on the distance between} {the gripper of the robot arm and the target position.}
\subsection{Proximal Policy Optimization}
{In this work, proximal policy optimization (PPO)~\cite{schulman2017proximal} is selected as our base RL algorithm. This is a state-of-the-art, on-policy actor-critic approach to training. The actor-critic architecture is common in deep RL; it is composed of an actor network which is used to output a policy, and a critic network which outputs a value to evaluate the current state, $s_{t}$. Proximal policy optimization (PPO) has been widely tested in robot control~\cite{andrychowicz2020learning} and video games~\cite{berner2019dota}. In contrast with the ``vanilla policy'' gradient algorithms, proximal policy optimization (PPO) learns the policy using a surrogate objective function
\begin{equation}
\centering
\mathcal{L}_{policy} = \mathbb{E}_{t}\biggr[\min \left (\frac{\pi_{\theta}(a_{t}|s_{t})}{\pi_{\theta_{old}}(a_{t}|s_{t})}A_{t}, \clip \left(\frac{\pi_{\theta}(a_{t}|s_{t})}{\pi_{\theta_{old}}(a_{t}|s_{t})}, 1-\epsilon, 1+\epsilon \right )A_{t}\right )\biggr],
\end{equation}
where $\pi_{\theta}(a_{t}|s_{t})$ is the current policy and $\pi_{\theta_{old}}(a_{t}|s_{t})$ is the old policy; $\epsilon$ is a clipping ratio which limits the change between the updated and the previous policy during the training process. $A_{t}$ is the advantage value which can be estimated as $R(s_{t}, a_{t}) - V(s_{t})$, with $R(s_{t}, a_{t})$ being the return value, and~$V(s_{t})$ the state value predicted by the critic network.}
\subsection{Hindsight Experiences and Goals}
The experiments follow the terminology suggested by OpenAI~\cite{plappert2018multi}, in which the possible goals are drawn from $\mathcal{G}$, and the goal being pursued does not influence the environment dynamics. In ESIL, two types of goal are recognised. One is the desired goal $g\in \mathcal{G}$, which is the target position {or state}, and may be different for different episodes. Within a single episode, $g$ is constant.  The second type of goal is the achieved goal $g^{ac}$, which is the achieved state in the environment, and this is considered to be different at each time step in an episode. In an episode, each transition can be represented as $(s_t|\langle g, g^{ac}_{t} \rangle, a_t, r_t, s_{t+1}|\langle g, g^{ac}_{t+1} \rangle)$, where $s_t$ indicates a state, $a_t$ indicates an action and $r_t$ indicates a~reward; $\langle \, , \rangle$ is simply used to represent grouping of goals.

In sparse reward settings, an agent will only get positive rewards when the desired goal $g$ is achieved. The sparse reward function can be defined as
\begin{equation}
r_{t}\left(g^{ac}_{t}, g\right):=
\begin{cases}
0& \text{if $\left\|g^{ac}_{t} - g\right\|\leq\epsilon$}\\
-1& \text{otherwise},
\end{cases}
\label{eq:sparse_reward}
\end{equation}
where $\epsilon$ is a threshold value, used to identify if the agent has achieved the goal. {However, the desired goal, $g$, might be difficult to reach during training. Thus, hindsight experiences are created through replacing the original desired goal $g$ with the current achieved goal $g_{t}^{ac}$ to augment the successful samples, and then reward $r_{t}$ can be recomputed according to Equation~\ref{eq:sparse_reward}. The modification of the desired goal can be denoted as $g^\prime$ and transitions from hindsight experiences can be represented as $(s_t|\langle g^\prime, g^{ac}_{t} \rangle, a_t, r_t, s_{t+1}|\langle g^\prime, g^{ac}_{t+1} \rangle)$.} {Intuitively, introducing $g^\prime$ serves a useful purpose in the early stages of training; taking, for example, a robot reaching task, the agent has no prior concept of how to move its effector to a specific location in space. Thus, even these original failed episodes contain valuable information for ultimately learning a useful control policy for the original, desired goal $g$.}

\section{Methodology}
\label{sec:method}
{The proposed} method combines PPO and episodic self-imitation learning to maximally use hindsight experiences for exploration to improve learning. {Recent advantages in episodic backward update~\cite{lee2019sample} and hindsight experiences~\cite{andrychowicz2017hindsight} are also leveraged} to guide exploration for on-policy RL. 

\subsection{Episodic Self-Imitation Learning}
{The present method} aims to use episodic hindsight experiences to guide the exploration of the PPO algorithm. To this end, {hindsight experiences are created from current episodes.} For an episode $i$, let there be $T$ time steps; after $T$, a series of transitions $\tau_i =\{ (s_{t}|\langle g, g^{ac}_{t} \rangle, a_{t}, r_{t}, s_{t+1}|\langle g, g^{ac}_{t+1}\rangle )\}_{t=0:T-1}$ is collected. If at time step $t=T-1$, in $s_{T}$, $g^{ac}_{T} \neq g$, it implies that in this episode, the agent failed to achieve the original goal. Simply, to create hindsight experiences, {the achieved goal $g^{ac}_{T}$ in the last state $s_{T}$ is selected and considered as the modified desired goal $g^{\prime}$, i.e., $g^\prime = g^{ac}_{T}$.} Next, a new reward $r_{t}^{\prime}$ is computed under the new goal $g^\prime$. Then, {a new ``imagined'' episode is achieved}, and a new series of transitions $\tau_i^{\prime} = \{(s_{t}|\langle g^\prime, g^{ac}_{t} \rangle, a_{t}, r_{t}^{\prime},s_{t+1}| \langle g^\prime, g^{ac}_{t+1}\rangle )\}_{t=0:T-1}$ is collected.

Then, an approach to self-imitation learning based on episodic hindsight experiences is proposed, which applies the policy updates to both hindsight and in-environment episodes. {Proximal policy optimization (PPO) is used as the base RL algorithm}, which is a state-of-the-art on-policy RL algorithm. 
With current and corresponding hindsight experiences, {a new objective function is introduced and defined as}
\begin{equation}
\centering
\mathcal{L} = \alpha\cdot\mathcal{L}_{PPO} + \beta\cdot\mathcal{L}_{ESIL},
\label{eq:loss}
\end{equation}
where $\alpha$ is the weight coefficient of $\mathcal{L}_{PPO}$. In the experiments, we set $\alpha=1$ as default {to balance the contribution of $\mathcal{L}_{PPO}$ and $\mathcal{L}_{ESIL}$}. $\mathcal{L}_{PPO}$ is the loss of PPO which can be written as
\begin{equation}
    \mathcal{L}_{PPO} = \mathcal{L}_{policy} - c\cdot \mathcal{L}_{value}
\label{eq:ppo_loss}
\end{equation}
{where $\mathcal{L}_{policy}$ is the policy loss which is parameterised by $\theta$, $\mathcal{L}_{value}$ is the value loss which is parameterised by $\eta$, and $c$ is the weight coefficient of the $\mathcal{L}_{value}$, which is set to 1 to match the default PPO setting~\cite{schulman2017proximal}. The policy loss, $\mathcal{L}_{policy}$, can be represented as}
\begin{equation}
\centering
\mathcal{L}_{policy} = \mathbb{E}_{s_{t}, a_{t}, g\in\Tau}\biggr[\min \left (\frac{\pi_{\theta}(a_{t}|s_{t}, g)}{\pi_{\theta_{old}}(a_{t}|s_{t}, g)}A_{t}, \clip \left (\frac{\pi_{\theta}(a_{t}|s_{t}, g)}{\pi_{\theta_{old}}(a_{t}|s_{t}, g)}, 1-\epsilon, 1+\epsilon \right )A_{t}\right )\biggr],
\label{eq:policy_loss}
\end{equation}
here, $A_{t}$ is the advantage value, and can be computed as $R_{t} - V_{\eta}(s_{t}, g)$. $V_{\eta}(s_{t}, g)$ is the state value at time step $t$ which is predicted by the critic network. $R_{t}$ is the return at time step $t$. $\epsilon$ is the clip ratio. $\Tau$~indicates original trajectories. {The value loss is an squared error loss $\mathcal{L}_{value}=(V_{\eta}(s_{t}, g) - R_{t})^{2}$.}

{For the $\mathcal{L}_{ESIL}$ term}, $\beta$ is an adaptive weight coefficient of $L_{ESIL}$; it can be defined as the ratio of samples which are selected for self-imitation learning
\begin{equation}
    \beta=\frac{N_{ESIL}}{N_{Total}},
\end{equation}
where $N_{ESIL}$ is the number of samples used for self-imitation learning and $N_{Total}$ is the total number of collected samples.
The episodic self-imitation learning loss $\mathcal{L}_{ESIL}$ can be written as 
\begin{equation}
\centering
\mathcal{L}_{ESIL}=\mathbb{E}_{s_{t}, a_{t}, g^\prime\in\Tau^{'}}\left[\log\pi_{\theta}\left(a_{t}|s_{t}, g^\prime\right)\cdot\mathcal{F}_t\right],
\label{eq:imitaion}
\end{equation}
where $\Tau^{'}$ indicates hindsight trajectories and $\mathcal{F}_t$ is the trajectory selection module which is based on returns of the current episodes, $R$, and the returns of corresponding hindsight experiences, $R^{\prime}$.
%indicates hindsight goal selection.
% Note that after each episode, we augment the original experiences by concatenating their hindsight experiences. After a batch of new transitions is collected, optimization is performed with minibatch stochastic gradient descent to maximize the objective.

\subsection{Episodic Update with Hindsight}%{Hindsight goal selection}
Two important issues of ESIL are: (1) hindsight experiences are sub-optimal, and (2) the {detrimental} effect of updating networks with correlated trajectories. Although episodic self-imitation learning makes exploration more effective, hindsight experiences are not from experts and not ``perfect'' demonstrations.  With the training process continuing, if the agent is always learning these imperfect demonstrations, the policy will be stuck at the sub-optimal, or experience overfitting.

\begin{figure}[t!]
\centering
\includegraphics[width=\columnwidth]{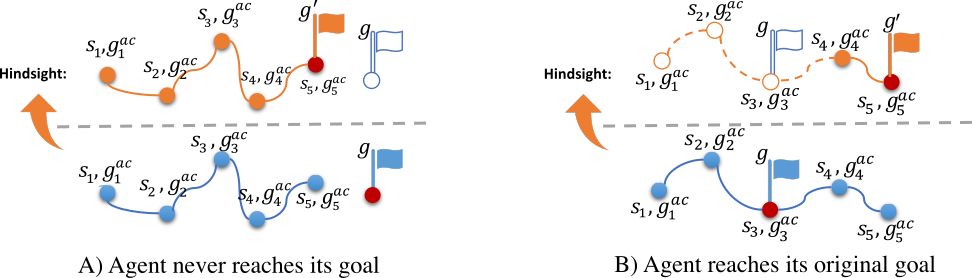}
\caption{A simplified illustration %Please change the ''A)'' and ''B)'' with ''(a)'',''(b)'' in the picture
of trajectory selection. Blue trajectories indicate original experiences. Orange trajectories indicate hindsight experiences. Solid trajectories in the hindsight experiences are selected by the trajectory selection module of ESIL with new ``imagined'' goals.}
\label{fig:hs_princple}
\end{figure}

To prevent the agent learning from imperfect hindsight experiences, {hindsight experiences are actively selected based on returns.} With the same action, different goals may lead to different results. The proposed method only selects hindsight experiences that can achieve higher returns. The illustration of the trajectory selection module is in Figure~\ref{fig:hs_princple}. For an episodic experience and its hindsight experience, the returns of the episodic experience and its hindsight experience can be calculated, respectively. 
In a trajectory, at time step $t$, the return $R_{t}$ can be calculated by $R_{t} = R(s_t, g) = r_{t} + \gamma \cdot r_{t+1} + \gamma^{2} \cdot r_{t+2} + ... + \gamma^{T-t-1} \cdot r_{T-1}$. Then, for a trajectory $\tau_{i}$, we have $\left\{R_{0}^{i}, R_{1}^{i}, R_{2}^{i}, ..., R_{T-1}^{i}\right\}$.
% \begin{figure}[h]
% \centering
% \includegraphics[width=\columnwidth]{figures/hgs.pdf}
% \caption{The brief illustration of trajectory selection. Blue trajectories indicate original experiences. Orange trajectories indicate hindsight experiences. Solid trajectories in the hindsight experiences are selected by the trajectory selection module of ESIL with new ``imagined'' goals.}
% \label{fig:hs_princple}
% \end{figure}
For the hindsight experiences, similarly, the return $R_{t}'$ for each time step, $t$,~with respect to the hindsight goals $g^\prime$, can be calculated. Based on the modified trajectory $\tau_{i}^{\prime}$ with the same length of $\tau_{i}$, we therefore have the returns $\left\{R_{0}^{i^\prime}, R_{1}^{i^\prime}, R_{2}^{i^\prime}, ..., R_{T-1}^{i^\prime}\right\}$. During training, the~hindsight experiences with higher returns are used for self-imitation learning. The rest of the hindsight experiences will be supposed to be worthless samples and ignored.
Then, Equation (\ref{eq:imitaion}) can be rewritten as
\begin{equation}
\centering
\mathcal{L}_{ESIL}=\mathbb{E}_{s_{t}, a_{t}, g^\prime\in\Tau^{'},g\in\Tau}\biggr[\log\pi_{\theta}\left(a_{t}|s_{t}, g^\prime\right)\cdot\mathcal{F}\left(s_{t}, g, g^\prime\right)\biggr],
\label{eq:final_loss}
\end{equation}
where $\mathcal{F}\left(s_{t}, g, g^{\prime}\right)$ is the trajectory selection module. The selection function can be expressed as
% \begin{equation}
% \centering
% \mathcal{F}\left(s_{t}, g^{de}, g^{de\prime}\right) = \min\left[\sgn \left(R^{\prime}(s_{t}, g^{de\prime}) - R(s_{t}, g^{de})\right), 0\right],
% \end{equation}
\begin{equation}
\centering
\mathcal{F}\left(s_{t}, g, g^\prime \right)=\mathds{1}\left[R(s_{t}, g^\prime) > R(s_{t}, g)\right],
\end{equation}
here, {$\mathds{1}(\cdot)$ is the unit step function}. Consider the OpenAI FetchReach environment as an example. For~a failed trajectory, the rewards $r_{t}$ are $\{-1,-1,\cdots, -1\}$. {The desired goal is modified} to construct a new hindsight trajectory and the new rewards $r_{t}^{\prime}$ become $\{-1,-1,\cdots, 0\}$. Then,  {$R$ and $R^{\prime}$ can be calculated~separately.}

From a goal perspective, episodic self-imitation learning (ESIL) tries to explore (desired) goals to get positive returns.~It can be viewed as a form of multi-task learning, {because ESIL has two objective functions to be optimised jointly.}~It is also related to self-imitation learning (SIL)~\cite{oh2018self}.~However,~ the~difference is that SIL uses $\left(R-V_{\theta}(s)\right)_{+}$ on past experiences to learn to choose the action chosen in the past in a given state, rather than goals. The full description of ESIL can be found in Algorithm~\ref{alg}.
%The selective procedure tends to use hindsight experiences that have better returns. At the early stage of training, there are a lot of failures. Then the selective procedure uses many hindsight experiences. At the late stage, there are also many successful experiences. Then the selective procedure compares the successful experiences and hindsight experiences and uses the better one. 
\begin{algorithm}[t!]
\caption{Proximal policy optimization (PPO) with Episodic Self-Imitation Learning (ESIL)}
\label{alg}
\begin{algorithmic}[1]
 \REQUIRE an actor network $\pi(s, g|\theta)$, a critic network $V(s, g|\eta)$, the maximum steps $T$ of an episode, a reward function $r$
 
\FOR{$\mathrm{iteration} = 1, 2, \cdots $}
\STATE $\Tau = \varnothing$, $\Tau^{\prime} = \varnothing$ 
\FOR{$\mathrm{episode} = 1, 2, ..., N$}
\STATE $\tau = \varnothing$
\FOR{$t = 0, 1, \cdots, T-1$}
\STATE Sample an action $a_{t}$ using the actor network $\pi(s_{t}, g|\theta)$
\STATE Execute the action $a_{t}$ and observe a new state $s_{t+1}$
\STATE Store the transition $\left(s_{t}|\langle g, g^{ac}_{t} \rangle, a_{t}, r_{t}, s_{t+1}|\langle g, g^{ac}_{t+1} \rangle \right)$ in $\tau$
\ENDFOR

\FOR{\textbf{each} transition $\left(s_{t}, a_{t}, r_{t}, g, g^{ac}_t \right)$ in $\tau$}
\STATE Clone the transition and replace $g$ with $g^{\prime}$, where $g^{\prime}=g^{ac}_{T}$
\STATE $r_{t}^{\prime}:=r\left(s_{t}, a_{t}, g^{\prime}\right)$
\STATE Store the transition $\left(s_{t}, a_{t}, r_{t}^{\prime}, g^{\prime}, g^{ac}_t\right)$ in $\tau^{\prime}$
\ENDFOR

\STATE Store the trajectory $\tau$ and the hindsight trajectory $\tau^{\prime}$ in $\Tau$ and $\Tau^{\prime}$, respectively 
%\STATE Store the hindsight trajectory $\tau^{\prime}$ in $\Tau^{\prime}$
\ENDFOR

\STATE Calculate the Return $R$ and $R^{\prime}$ for all transitions in $\Tau$ and $\Tau^{\prime}$, respectively
%\STATE Calculate the Return $R^{\prime}$ for all transitions in $\Tau^{\prime}$

\STATE Calculate the PPO loss: $\mathcal{L}_{PPO} = \mathcal{L}_{policy}(\theta) - c\cdot \mathcal{L}_{value}(\eta)$ using $\Tau$ \eqref{eq:ppo_loss}
\STATE Calculate the ESIL loss: $\mathcal{L}_{ESIL}(\theta)$ using $\Tau^{\prime}$, $R$ and $R^{\prime}$ \eqref{eq:imitaion}
%\STATE Calculate the critic loss: $\mathcal{L}_{c}(\theta^{c}) = \mathbb{E}_{s_{t}, g\in \Tau}\left[\left(R - v(s_{t}, g|\theta^{c})\right)^{2}\right]$ using $\Tau$ and $R$
\STATE Update the parameters $\theta$ and $\eta$ using loss $\mathcal{L}=\alpha\cdot\mathcal{L}_{PPO} + \beta\cdot\mathcal{L}_{ESIL}$ \eqref{eq:loss}
%\STATE Update the parameters $\theta^{c}$ using loss $\mathcal{L}_{c}$
\ENDFOR 
\end{algorithmic}
\end{algorithm}

\section{Experiments and Results}
\label{sec:experiment}
\begin{figure}[t]
\minipage{0.19\textwidth}
  \centering
  \includegraphics[width=\linewidth]{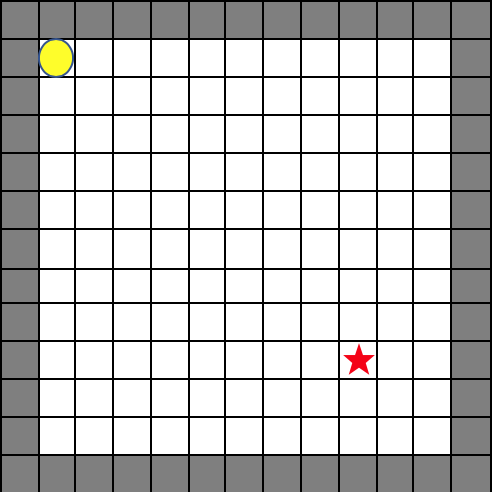}
  ({a}) Empty Room%\hspace{3em}          
\endminipage\hfill
\minipage{0.19\textwidth}
  \centering
  \includegraphics[width=\linewidth]{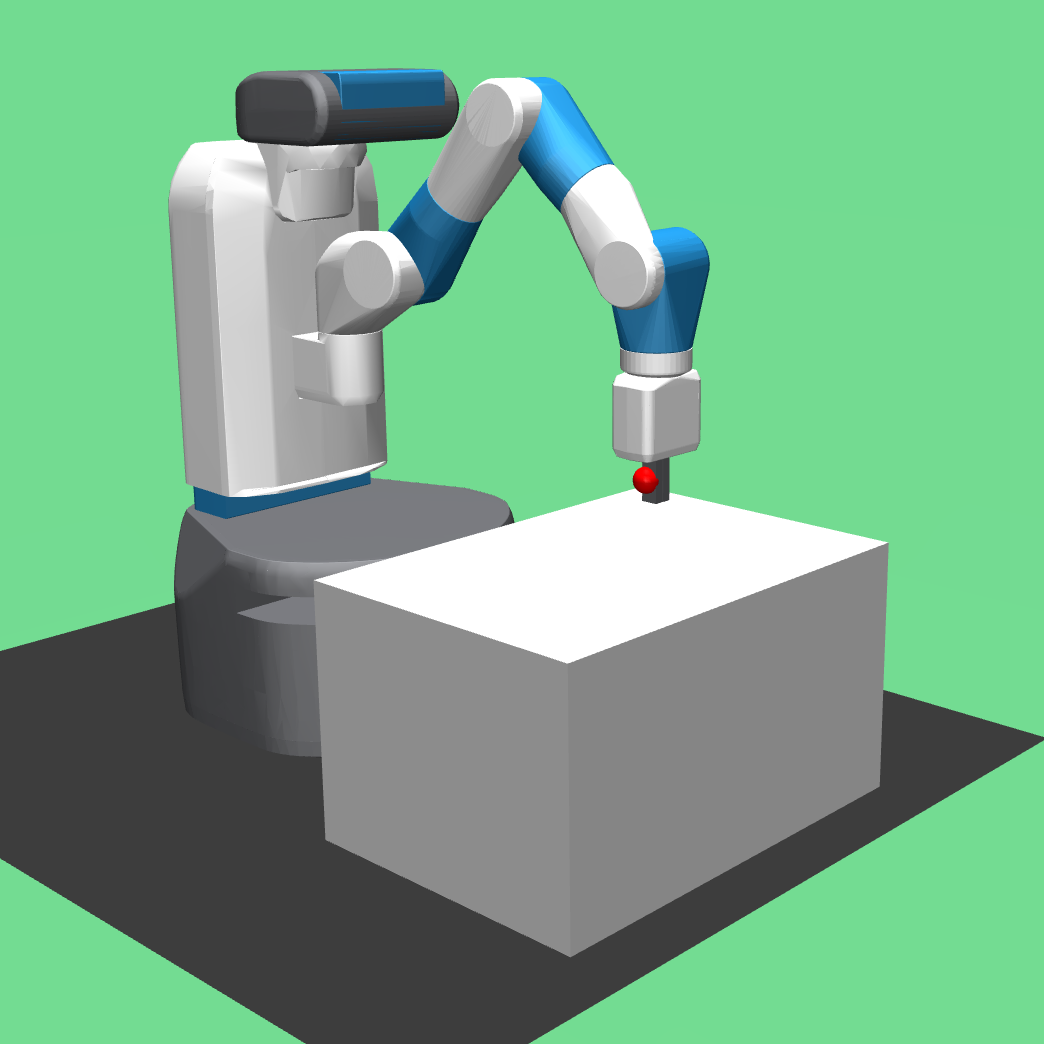}
  ({b}) FetchReach%\hspace{3em}  
\endminipage\hfill
\minipage{0.19\textwidth}%
  \centering
  \includegraphics[width=\linewidth]{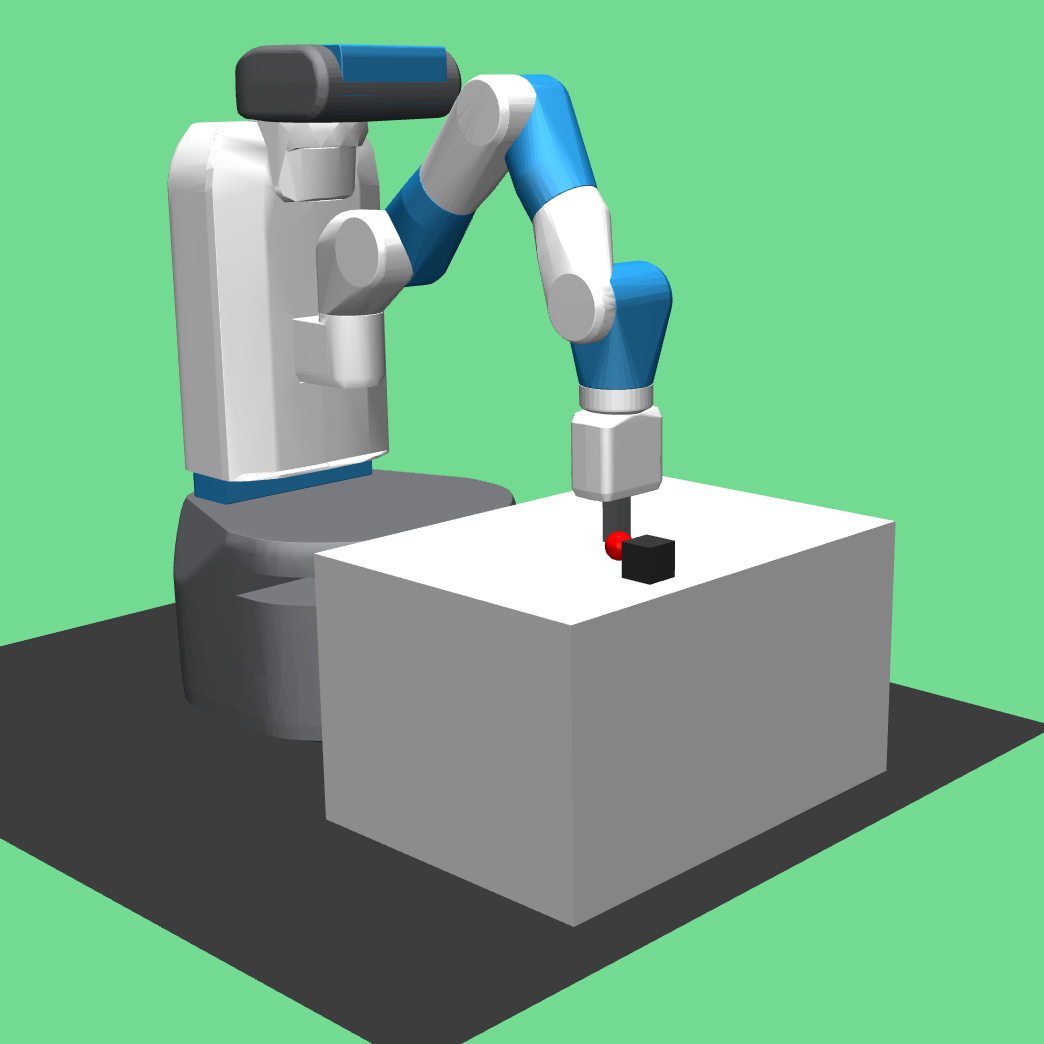}
  ({c}) FetchPush%\hspace{3em}        
\endminipage\hfill
\minipage{0.19\textwidth}%
  \centering
  \includegraphics[width=\linewidth]{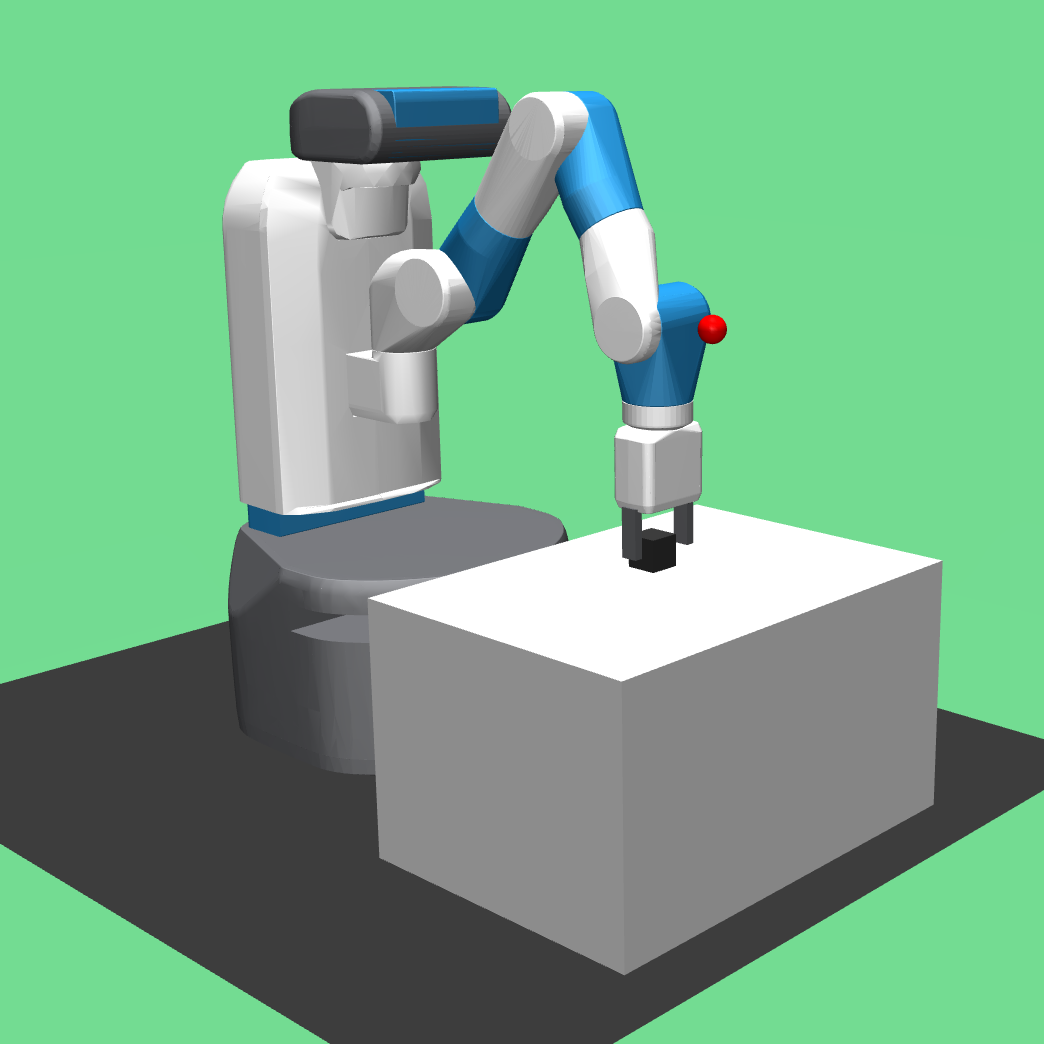}
  ({d}) FetchPickPlace
\endminipage\hfill
\minipage{0.19\textwidth}%
  \centering
  \includegraphics[width=\linewidth]{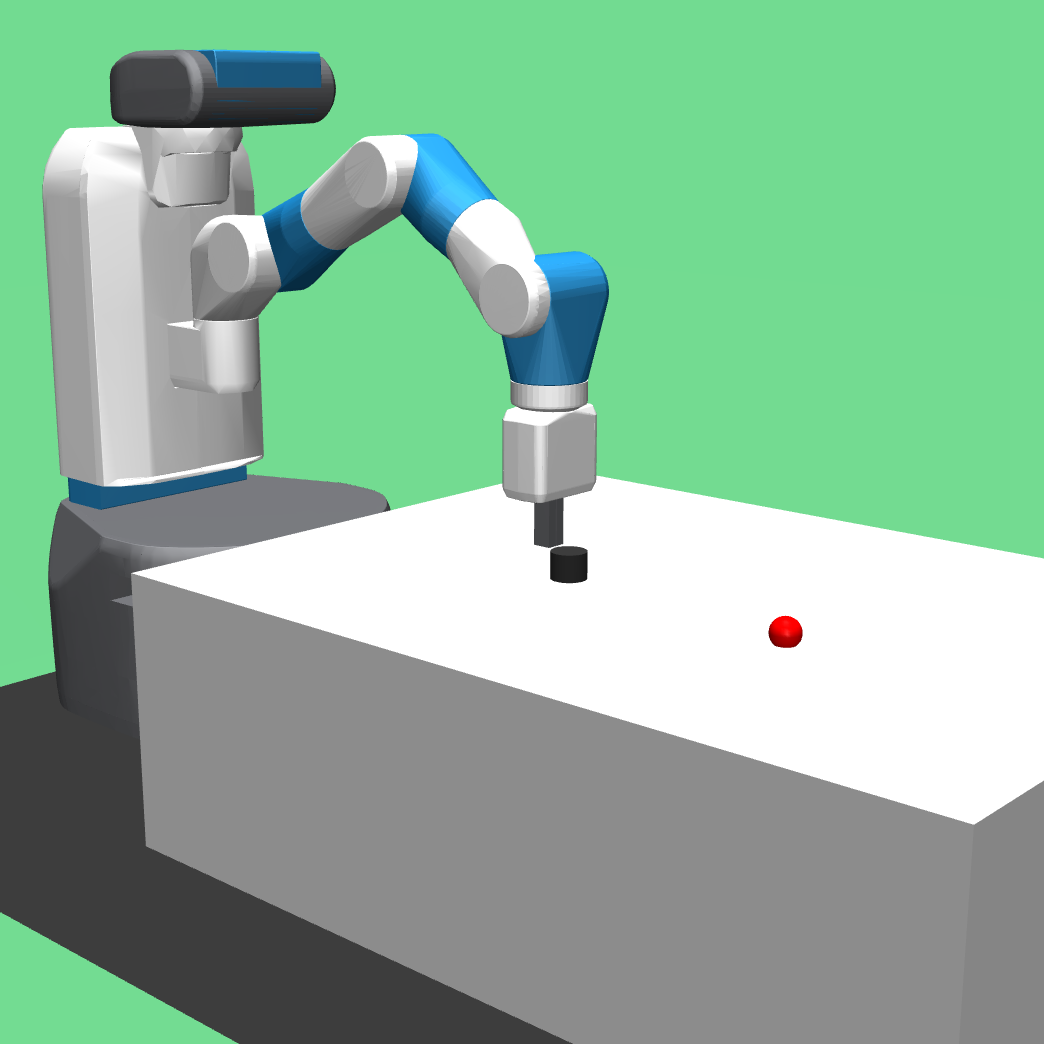}
  ({e}) FetchSlide %\hspace{3em}       
\endminipage
\caption{Evaluation environments. ({a}) is the Empty Room environment, in which a yellow circle indicates the position of the agent and a red star represents a target position. ({b}--{e}) are the Fetch robotic environments. The red spot represents a target position.}
\label{fig:env}
\end{figure}

{The proposed method is evaluated on several multi-goal environments}, including the Empty Room environment and the OpenAI Fetch environments (see Figure~\ref{fig:env}). The Empty Room environment is a toy example, and has discrete action spaces. In the Fetch environments, there are four robot tasks with continuous action spaces.
To obtain a comprehensive comparison between the proposed method and other baseline approaches, suitable baseline approaches are selected for different environments. 
{Ablation studies of the trajectory selection module are also performed}.

\subsection{Setup}

\textbf{{Empty Room (grid-world) environment}%Please confirm whether the bold format is  necessary? If not, please remove it.
}:
The Empty Room environment is a simple grid-world environment. The agent is placed in an $11\times11$ grid, representing the room. The goal of the agent is to reach a target position in the room. The start position of the agent is at the left upper corner of the room, and the target position is randomly selected  within the room. When the agent chooses an action that would lead it to fall outside the grid area, the agent stays at the current position. The length of each episode is 32. 
The desired goal, $g$, is a two-dimensional grid coordinate which represents the target position. The achieved goal, $g^{ac}_{t}$, is also a two-dimensional coordinate which represents the current position of the agent at time step $t$, and finally, the observation is a two-dimensional coordinate which represents the current position of the agent. 
The agent has five actions: left, right, up, down and stay; the agent executes a random action with probability 0.2. The agent can get $+1$ as a reward only when $g^{ac}_{t}=g$, otherwise, it gets a reward of $0$. 

The agent is trained with 1 CPU core. In each epoch, 100 episodes are collected for the training. After each epoch, the agent is evaluated for 10 episodes. During training, the actions are sampled from the categorical distribution. During evaluation, the action with the highest probability will be chosen.

\textbf{{Fetch robotic (continuous) environments}%.
}~\cite{plappert2018multi}:~The Fetch robotic environments are physically plausible simulations based on the real Fetch robot. The purpose of these environments is to provide a platform to tackle problems which are close to practical challenging robot manipulation tasks. Fetch is a 7-DoF robot arm with a two finger gripper. The Fetch environments include four~tasks: \emph{FetchReach}, \emph{FetchPush}, \emph{FetchPickAndPlace} and \emph{FetchSlide}. 
For all Fetch tasks, the length of each episode is 50. The~desired goal, $g$, is a three-dimensional coordinate which represents the target position. If~a task has an object, the achieved goal $g^{ac}_{t}$ is a three-dimensional coordinate represents the position of the object. Otherwise, $g^{ac}_{t}$ is a three-dimensional coordinate represents the position of the gripper.
Observations include the following information: position, velocity and state of the gripper. If a task has an object, the position, velocity and rotation information of the object is included. Therefore, the~observation of FetchReach is a 10-dimensional vector.~The observation of other tasks is a 25-dimensional vector.~The~action is a four-dimensional vector. The first three dimensions represent the relative position that the gripper needs to move in the next step. The last dimension indicates the distance between the fingers of the gripper.
The reward function can be written as $r_{t}=-\mathds{1}\left(\left\|g^{ac}_{t} - g\right\|>\epsilon\right)$, where $\epsilon =0.05$.

In the Fetch environments, for FetchReach, FetchPush and FetchPickAndPlace tasks, the agent is trained using 16 CPU cores. In each epoch, 50 episodes are collected for training. The FetchSlide task is more complex, so 32 CPU cores are used. In each epoch, 100 episodes are collected for training. The~{Message Passing Interface (MPI) framework is used to perform}  synchronization when updating the network. After each epoch, the agent is evaluated for 10 episodes by each MPI worker. Finally,~the~success rate of each MPI worker is averaged. During training, actions are sampled from {multivariate normal} distributions. In the evaluation phase, the mean vector of the distribution is used as an action. 

{The proposed method, termed PPO+ESIL, is compared} with different baselines on different environments. All experiments are plotted based on five runs with different seeds. The solid line is the median value. The upper bound is the {75th} percentile and the lower bound is the {25th} percentile.

\subsection{Network Structure and Hyperparameters}

\textbf{{Network structure}%Please confirm whether the bold format is  necessary? If not, please remove it.
}: Both the actor network and the critic network have three hidden layers with 256~neurons. ReLu is selected as the activation function for the hidden layers. In the grid-world environment, the actor network builds a categorical distribution. In the Fetch environment, the actor network builds normal {distributions by producing mean vectors and the standard deviations of the independent variables}.

\textbf{{Hyperparameters}%.Please confirm whether the bold format is  necessary? If not, please remove it.
}: For all experiments, the learning rate is 0.0003 for both the actor and critic networks. The discount factor $\gamma$ is 0.98. Adam is chosen as an optimiser with $\epsilon=0.00001$. For each epoch, the~actor network and critic network are updated 10 times. The clip ratio of the PPO algorithm is 0.2. For the grid-world environment, it trains networks for 100 epochs with batch size equals 160. Each~epoch consists of 100 episodes. For the Fetch environments, in FetchReach task, it trains networks for 100 epochs and other tasks for 1000 epochs with batch size equals to 125. For FetchReach, FetchPush and FetchPickAndPlace tasks, each epoch consists of 50 episodes. For FetchSlide task, each epoch consists of 100 episodes. {In designing the experiments, the number of episodes within an epoch is a balance between being able to train, the length of time required to run experiments and the maximum number of time steps that would be required to a achieve a goal. All environments have a fixed maximum number of time-steps , but this maximum differs depending on the problem or environment. This means that the number of state--action pairs can differ between two environments that have the same number of episodes and the same number of epochs. We arrange the episodes to try to compensate for the number of state--action pairs collected during training to make experiments easier to compare.} {The models are trained on a machine with an Intel i7-5960X CPU and 64GB RAM.} 

%\section{Results and Discussions}
\subsection{Grid-World Environments}
\begin{figure}[t]
\minipage{0.5\textwidth}
  \centering
  \includegraphics[width=\linewidth]{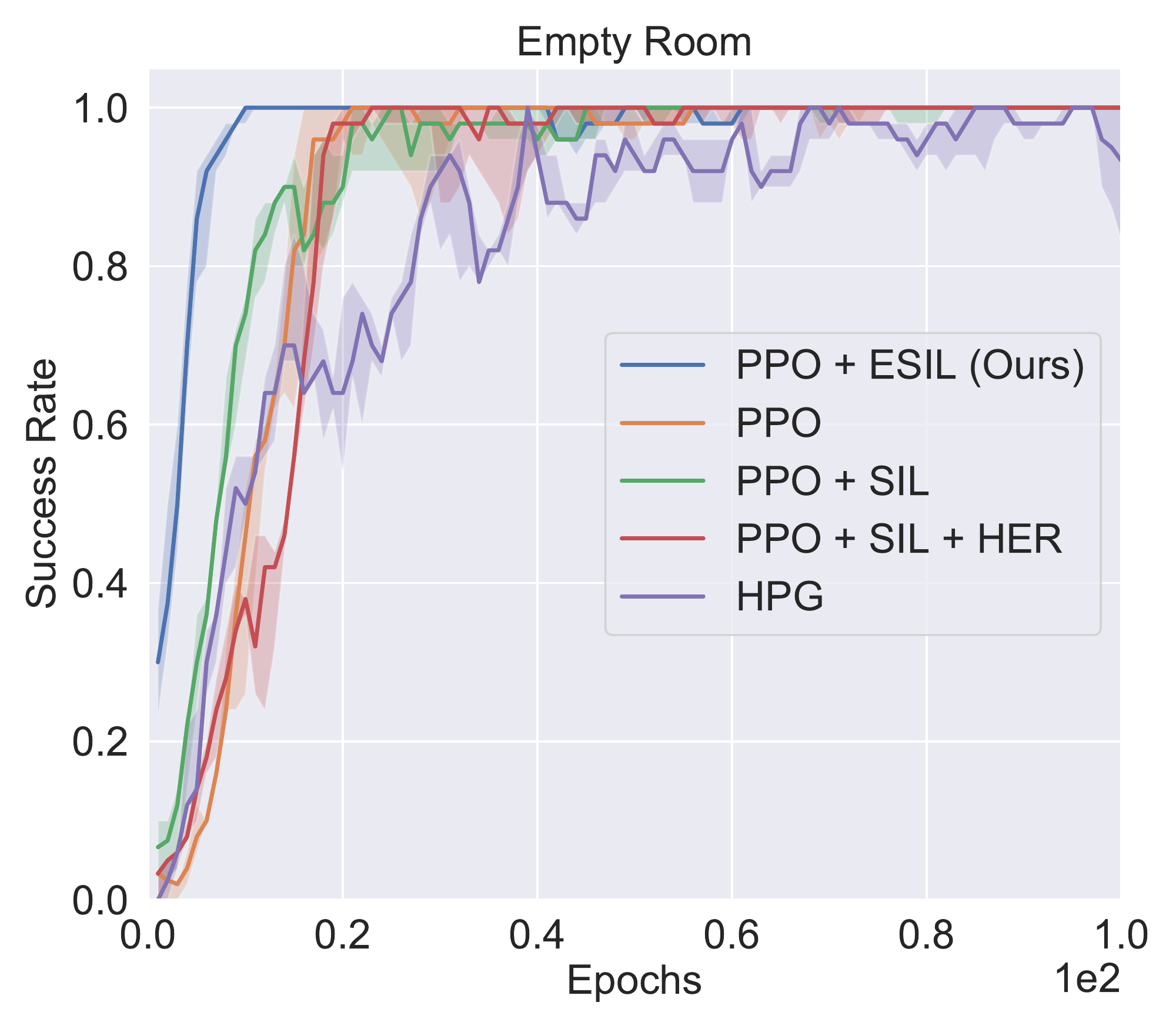}
  ({a}) Comparison with on-policy baselines\hspace{3em} 
\endminipage
\minipage{0.5\textwidth}
  \centering
  \includegraphics[width=\linewidth]{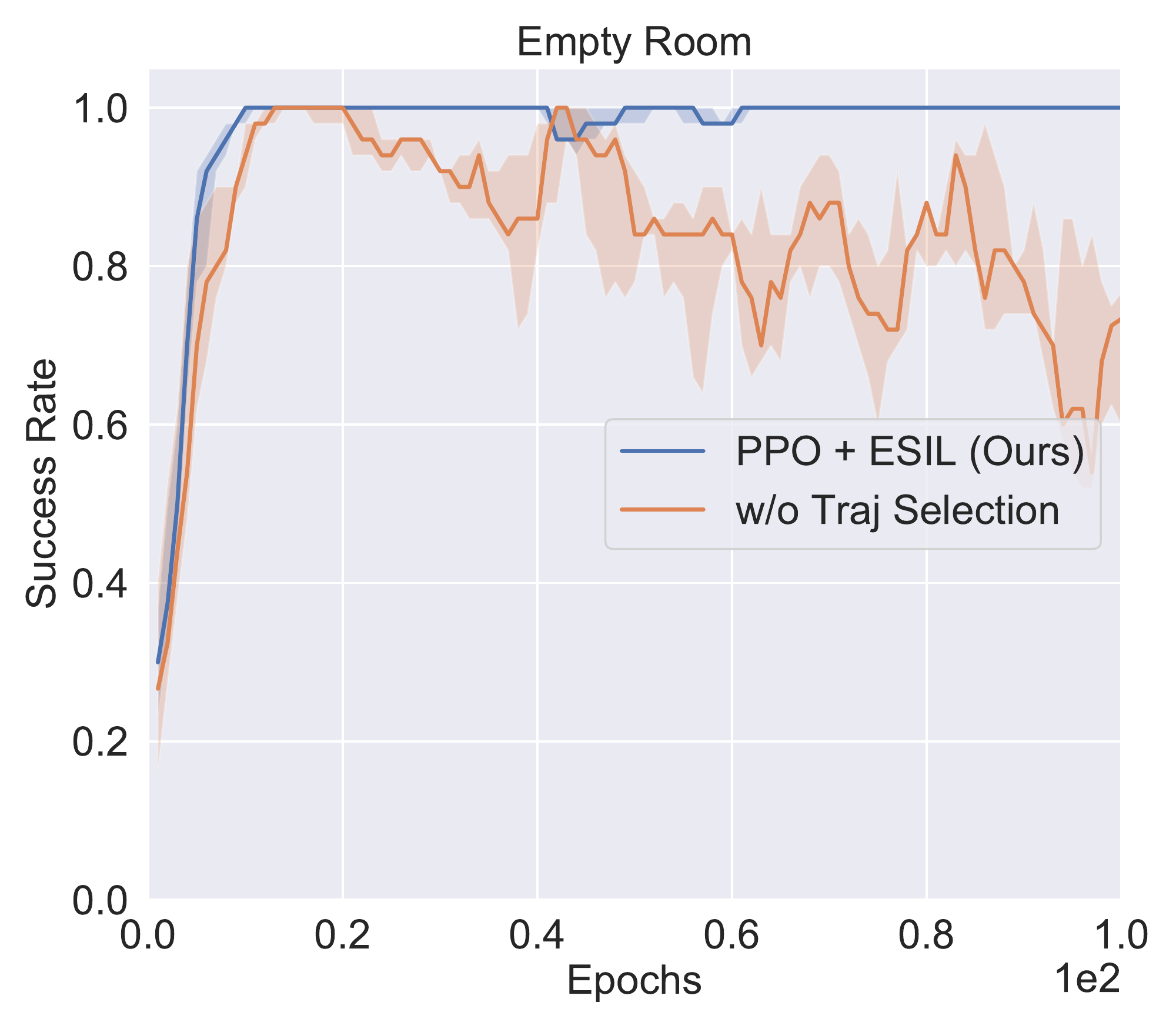}
  ({b}) {Ablation study of selection module}
\endminipage\hfill
\minipage{0.5\textwidth}%
  \centering
  \includegraphics[width=\linewidth]{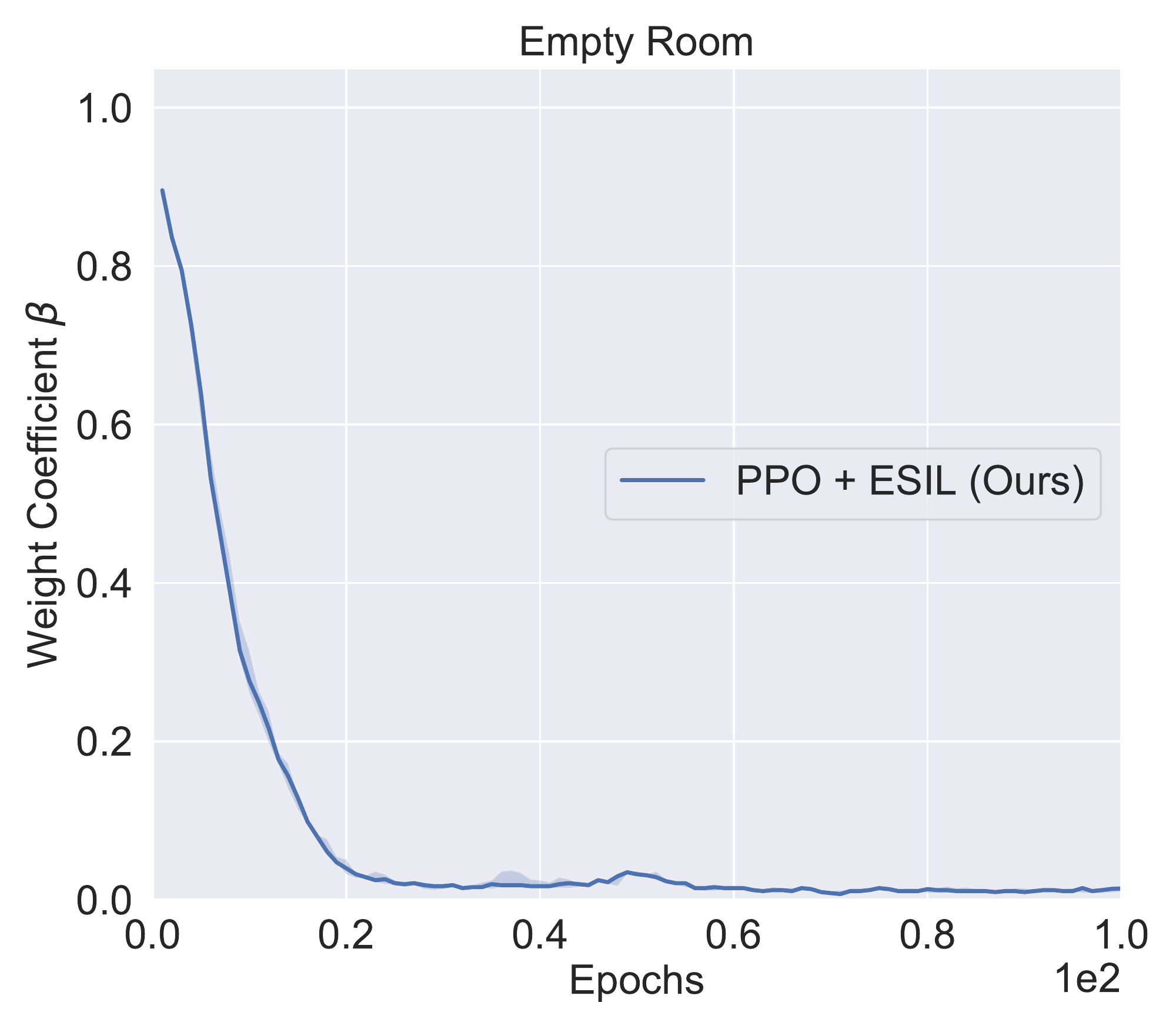}
  ({c}) Variation of adaptive weight coefficient $\beta$
\endminipage
\minipage{0.5\textwidth}%
  \centering
  \includegraphics[width=\linewidth]{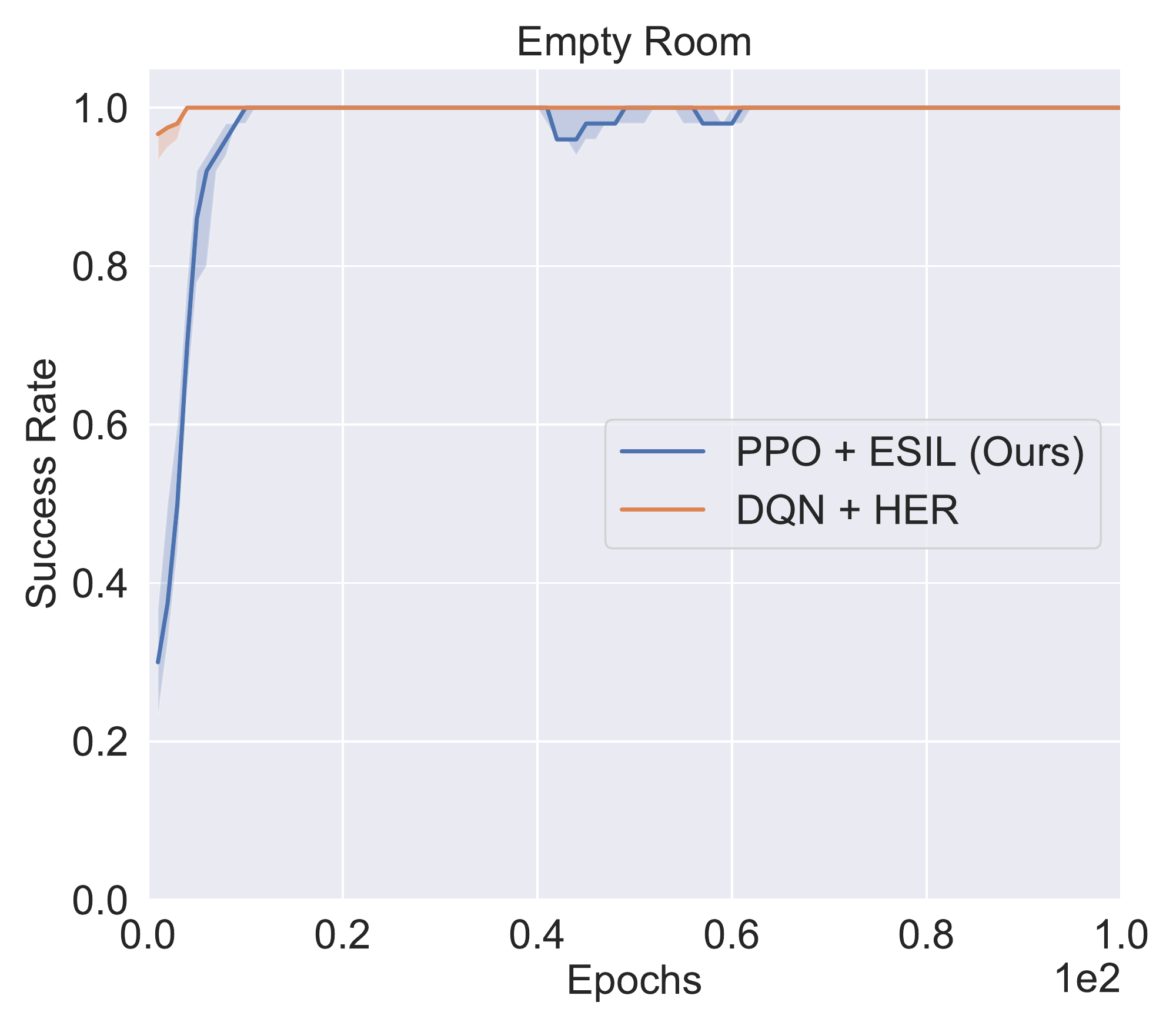}
  ({d}) Comparison with off-policy baselines
\endminipage\hfill
\caption{Results of the grid-world environment. ({a}) Comparing the performance of PPO+ESIL between the on-policy approaches. ({b}) An ablation study on the trajectory selection module. ({c}) The variation of adaptive weight coefficient $\beta$ through training. ({d}) Comparison of the performance of PPO+ESIL to an off-policy approach: DQN+HER.}
\label{fig:toy_results}
\end{figure}

To understand the basic properties of the proposed method, the toy Empty Room environment is used to evaluate ESIL.
%We compare the proposed method, named as PPO+HSL, with other baselines on the grid-world environment.
The following baselines are considered:
\begin{itemize}
    \item PPO: vanilla PPO~\cite{schulman2017proximal} for discrete action spaces;
    \item PPO+SIL/PPO+SIL+HER: Self-imitation learning (SIL) is used with PPO to solve hard exploration environments by imitating past good experiences~\cite{oh2018self}. In order to solve sparse rewards tasks, hindsight experience replay (HER) is applied to sampled transitions;
    \item DQN+HER: Hindsight experience replay (HER), designed for sparse reward problems, is~combined with a deep Q-learning network (DQN)~\cite{andrychowicz2017hindsight}; this is an off policy algorithm;
    \item Hindsight Policy Gradients (HPG): the vanilla implementation of HPG that is only suitable for discrete action spaces~\cite{rauber2018hindsight}.
\end{itemize}
More specifically, {PPO+ESIL is compared with above baseline methods} in Figure~\ref{fig:toy_results}a. This shows that PPO+ESIL converges faster than the other four baselines, and PPO+SIL converges faster than vanilla PPO, because PPO+SIL reuses past good experiences to help exploration and training. Hindsight~Policy Gradient (HPG) is slower than the others because goal sampling is not efficient and also unstable. 

Further, {the performance of the trajectory selection module is evaluated} in Figure~\ref{fig:toy_results}b. This shows that the selection strategy helps improve the performance. Hindsight experiences are not always perfect; the trajectory selection module filters some undesirable, modified experiences. Through adopting this selection strategy, the chance of agents learning from poor trajectories is reduced. The adaptive weight coefficient $\beta$ is also investigated in these experiments. {In Figure~\ref{fig:toy_results}c, it can be seen that at the initial stages of training, $\beta$ is high}. {This is because at this stage, the agent very seldom achieves the original goals}. The hindsight experiences can yield higher returns than the original experiences. {Therefore,~a~large proportion of hindsight experiences are selected to conduct self-imitation learning, helping the agent learn a policy for moving through the room}. In the later stages of training, the agent can achieve success frequently, and the hindsight experiences might be redundant (e.g., $R(s_{t}, g)\geq R(s_{t}, g^{\prime})$). In~this case, undesired hindsight experiences are removed by using the trajectory selection module and $L_{PPO}$ leads the training. However, when the trajectory selection module is not employed, all hindsight experiences are used through the entire training process which includes the redundant hindsight experiences. This leads to overfitting and makes training unstable. Thus, the $L_{ESIL}$ can provide the agent with a better initial policy, and the adaptive weight coefficient $\beta$ can balance the contributions of $L_{PPO}$ and $L_{ESIL}$ properly during training.
\begin{figure}[t]
\centering
\minipage{0.5\textwidth}
  \centering
  \includegraphics[width=\linewidth]{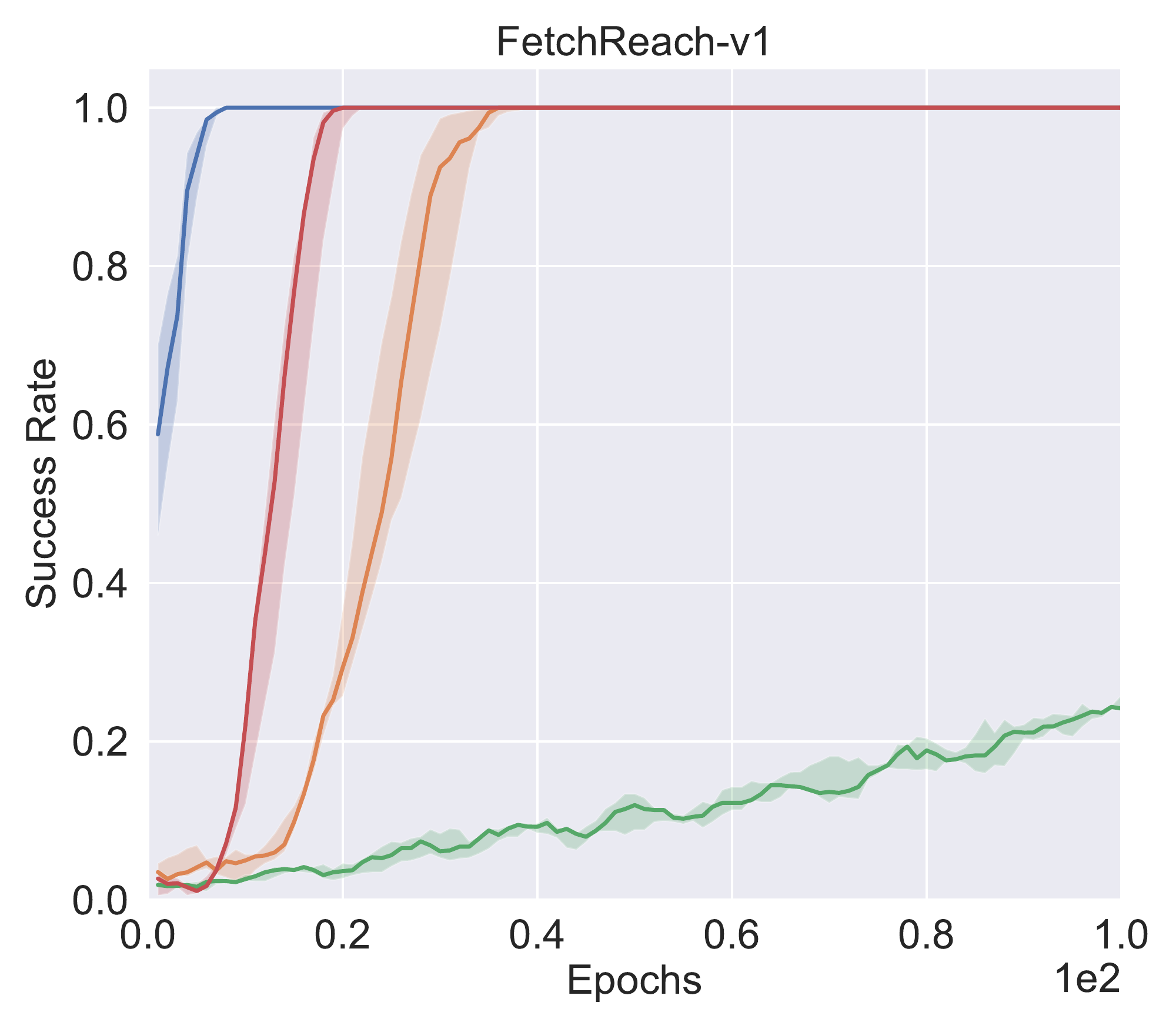}
  ({a}) FetchReach
\endminipage
\minipage{0.5\textwidth}%
  \centering
  \includegraphics[width=\linewidth]{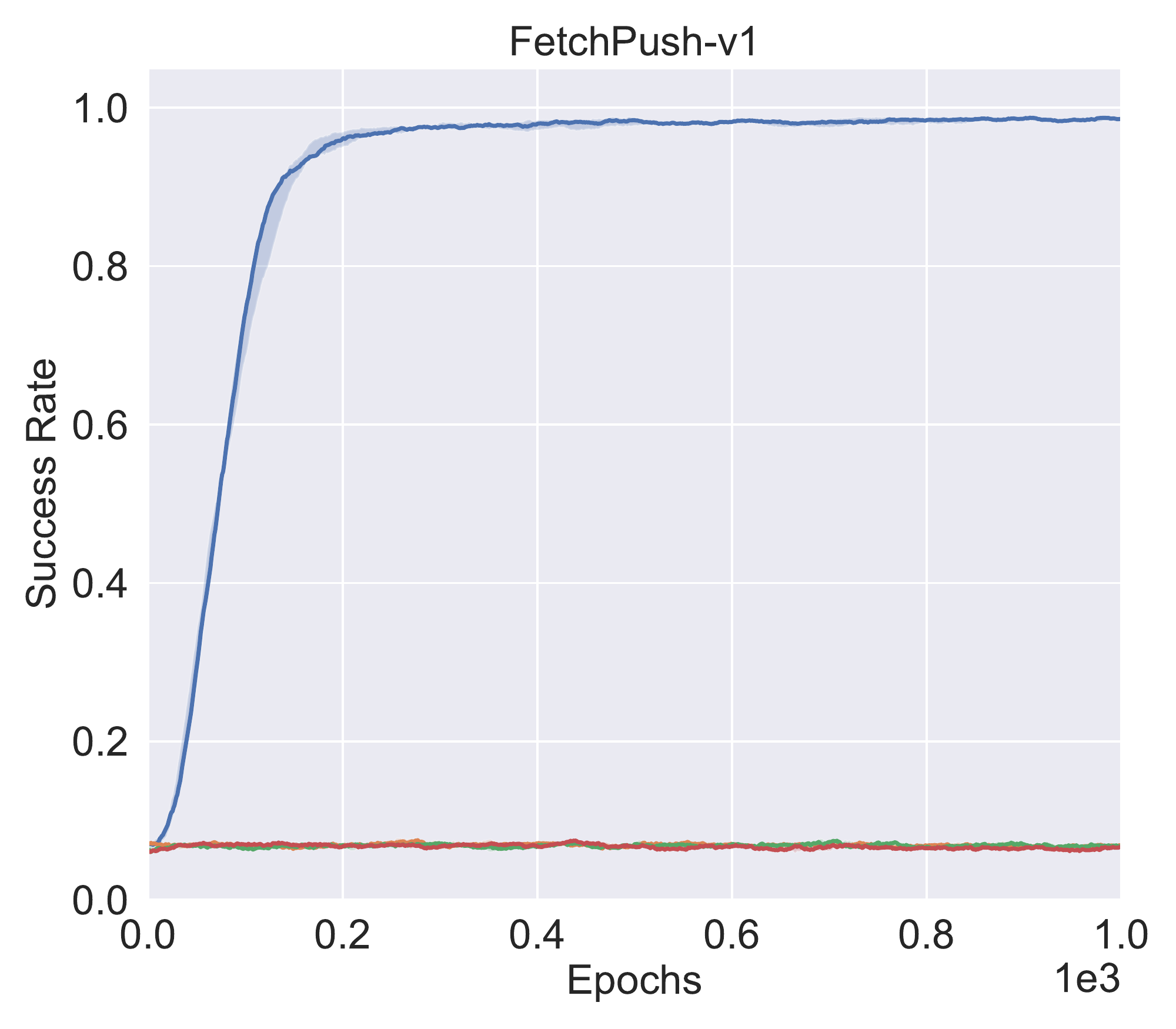}
  ({b}) FetchPush
\endminipage\hfill
\minipage{0.5\textwidth}%
  \centering
  \includegraphics[width=\linewidth]{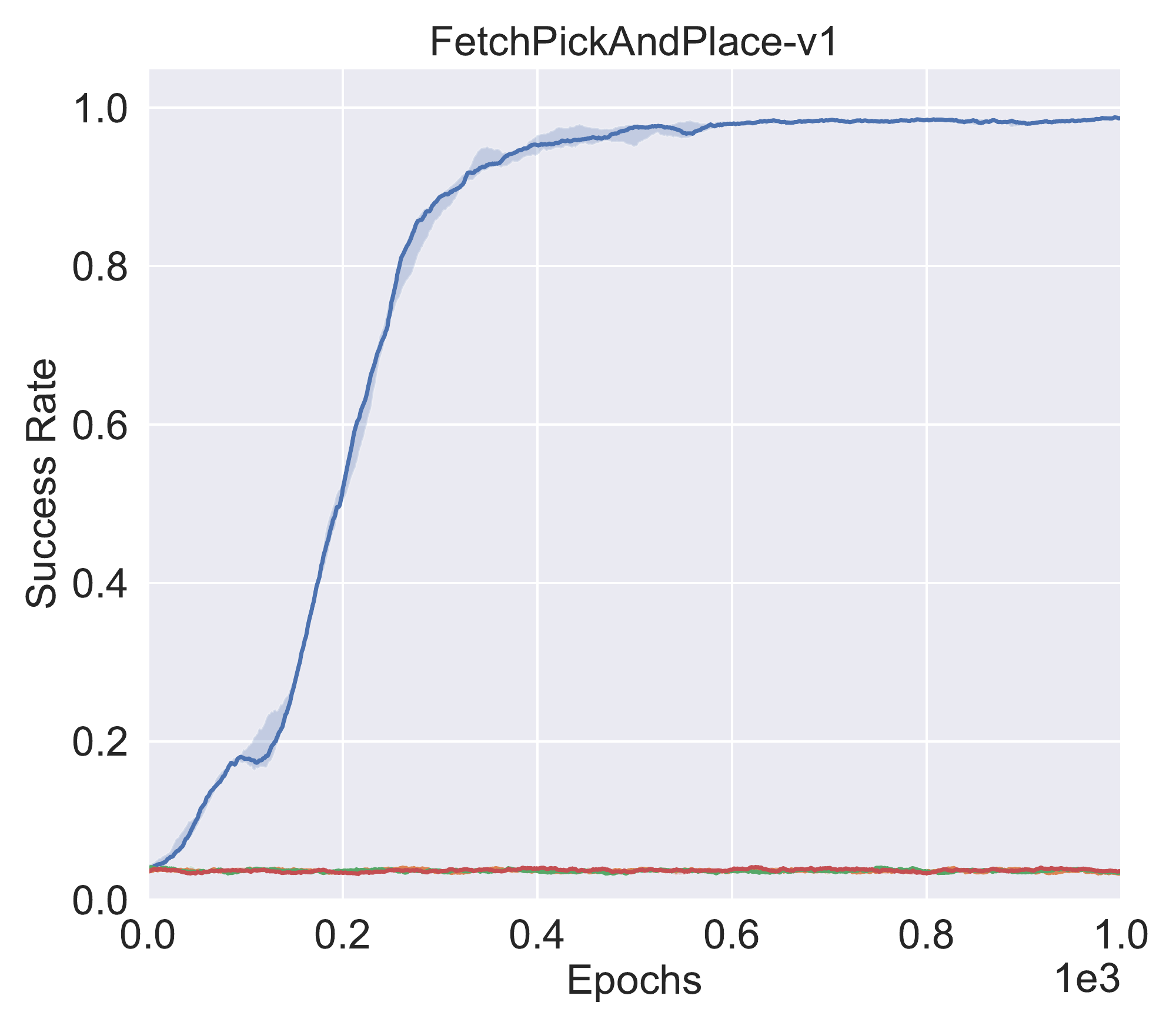}
  ({c}) FetchPickAndPlace
\endminipage
\minipage{0.5\textwidth}%
  \centering
  \includegraphics[width=\linewidth]{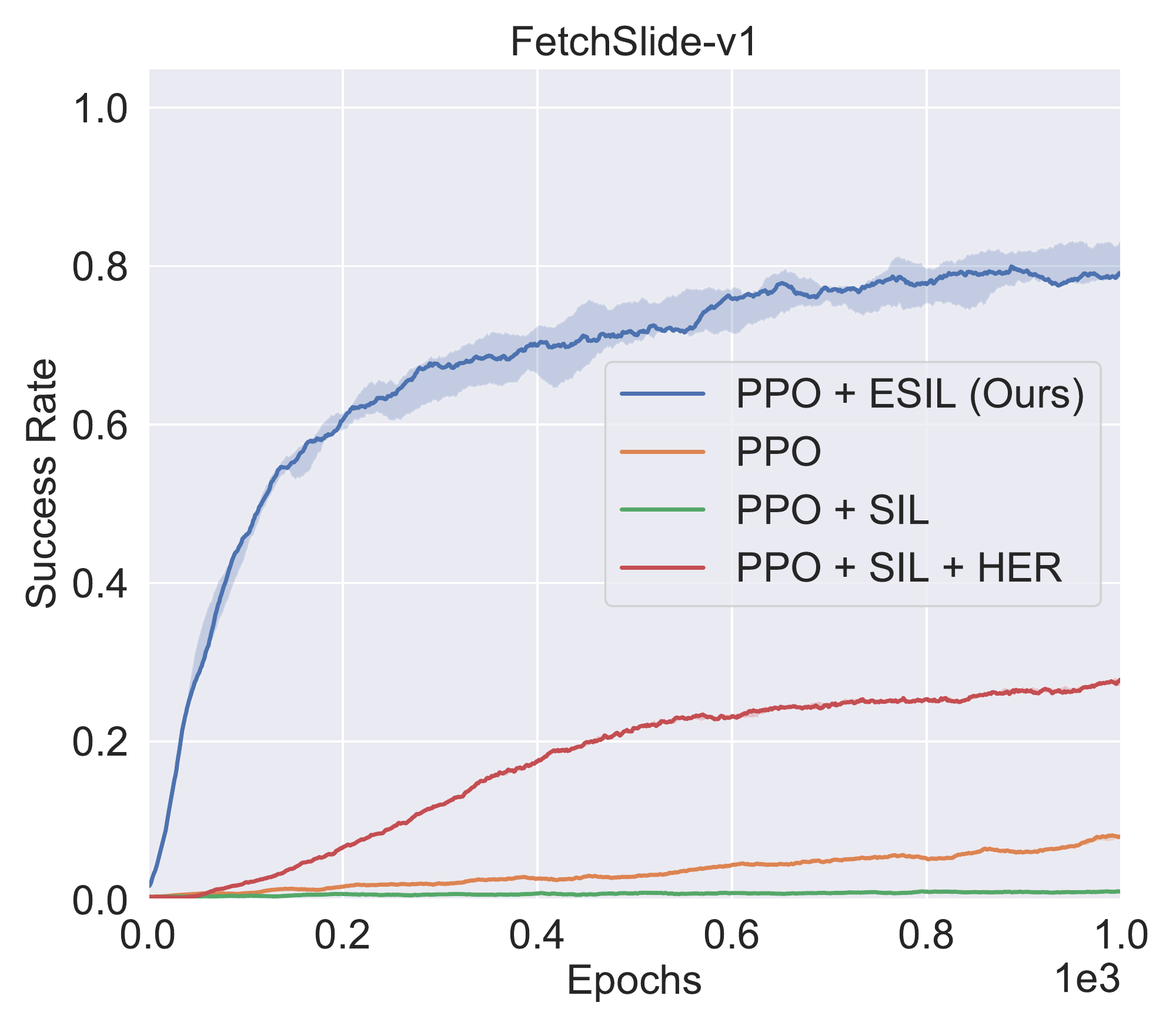}
  ({d}) FetchSlide
\endminipage\hfill
\caption{Results of comparison between ESIL and on-policy baselines on all Fetch environments.}
\label{fig:baseline_compare}
\end{figure}
Finally, {the combination of PPO+ESIL is also compared with DQN+HER,} which is an off-policy RL algorithm, in Figure~\ref{fig:toy_results}d. This shows that DQN+HER works a little better than ESIL at the {start of training}. However, the proposed method achieves similar results to DQN+HER later in training.

\subsection{Continuous Environments}
{Continuous control problems are generally more challenging for reinforcement learning. In the experiments of this section, the aim is to investigate how useful the proposed method is for several hard exploration OpenAI Gym Fetch tasks. These environments are commonly used to assess the performance of RL methods for continuous control.} The following baselines are considered: 
\begin{itemize}
    \item PPO: the vanilla PPO~\cite{schulman2017proximal} for continuous action spaces;
    \item PPO+SIL/PPO+SIL+HER: Self-imitation learning is used with PPO to solve hard exploration environments by imitating past good experiences~\cite{oh2018self}. For sparse rewards tasks, hindsight experience replay (HER) is applied to sampled transitions;
    \item DDPG+HER: this is the state-of-the-art off-policy RL algorithm for the Fetch tasks. Deep~deterministic policy gradient (DDPG) is trained with HER to deal with the sparse reward problem~\cite{andrychowicz2017hindsight}.
\end{itemize}
\begin{figure}[t]
\centering
\minipage{0.5\textwidth}
  \centering
  \includegraphics[width=\linewidth]{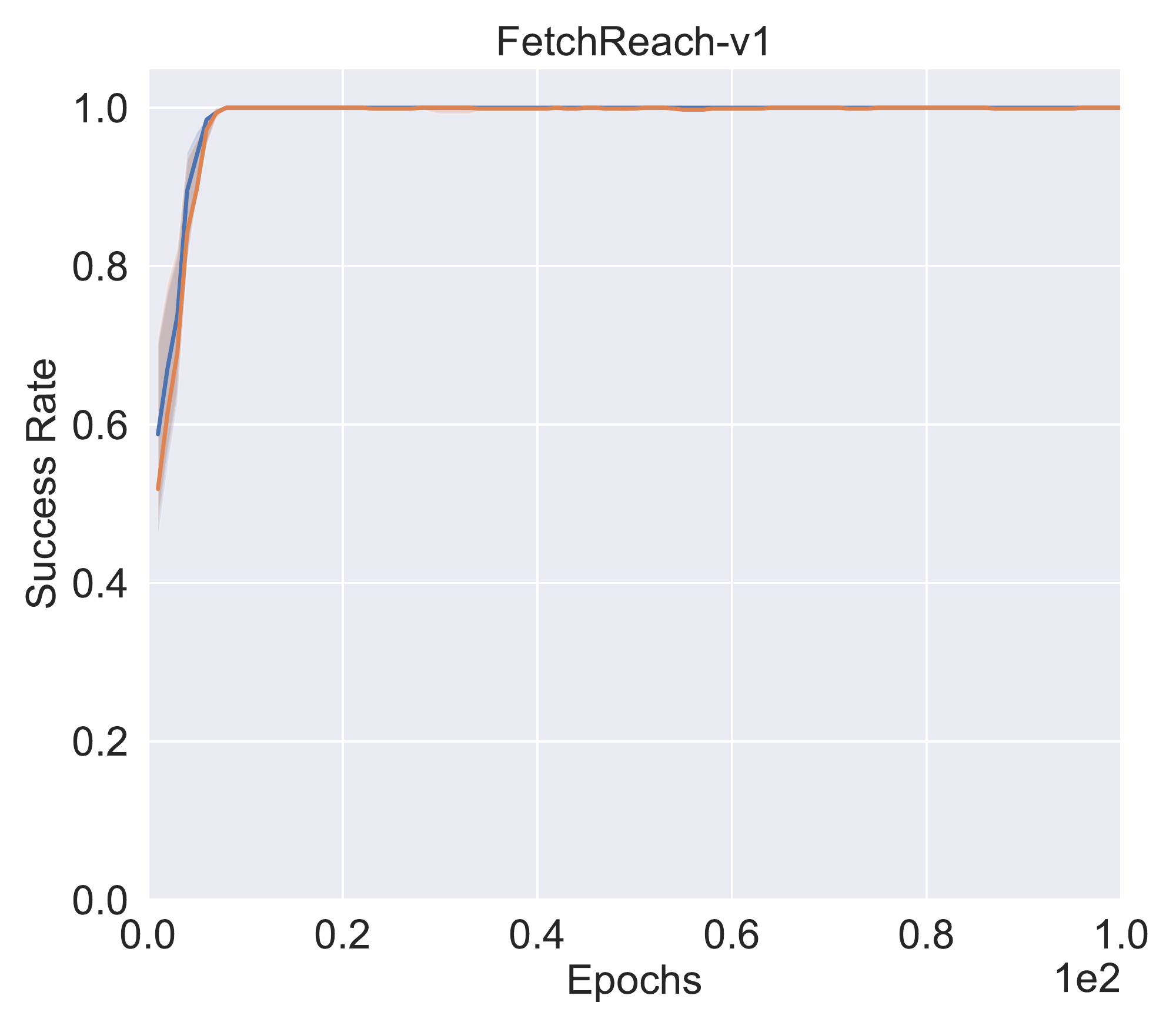}
  ({a}) FetchReach
\endminipage
\minipage{0.5\textwidth}%
  \centering
  \includegraphics[width=\linewidth]{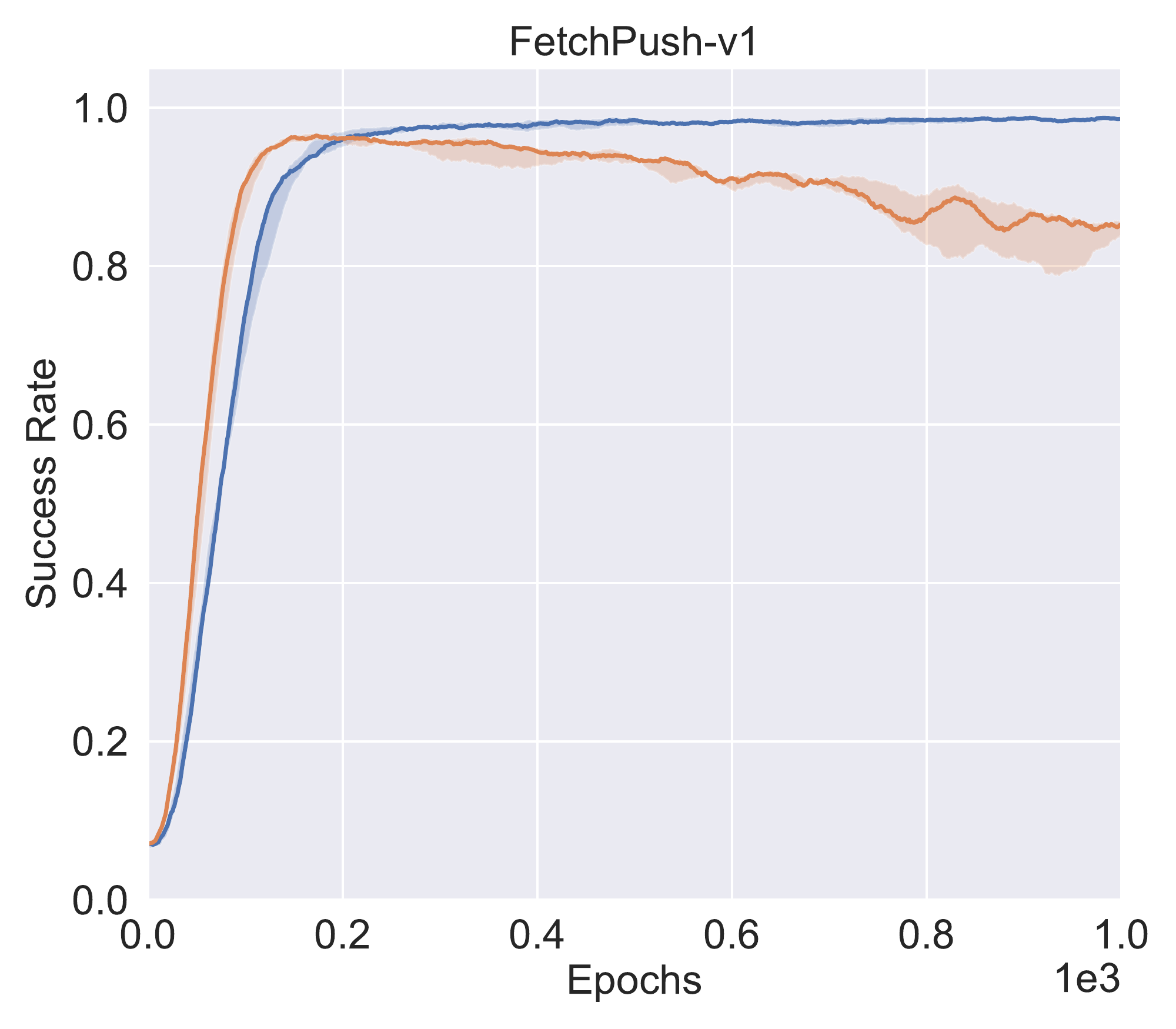}
  ({b}) FetchPush
\endminipage\hfill
\minipage{0.5\textwidth}%
  \centering
  \includegraphics[width=\linewidth]{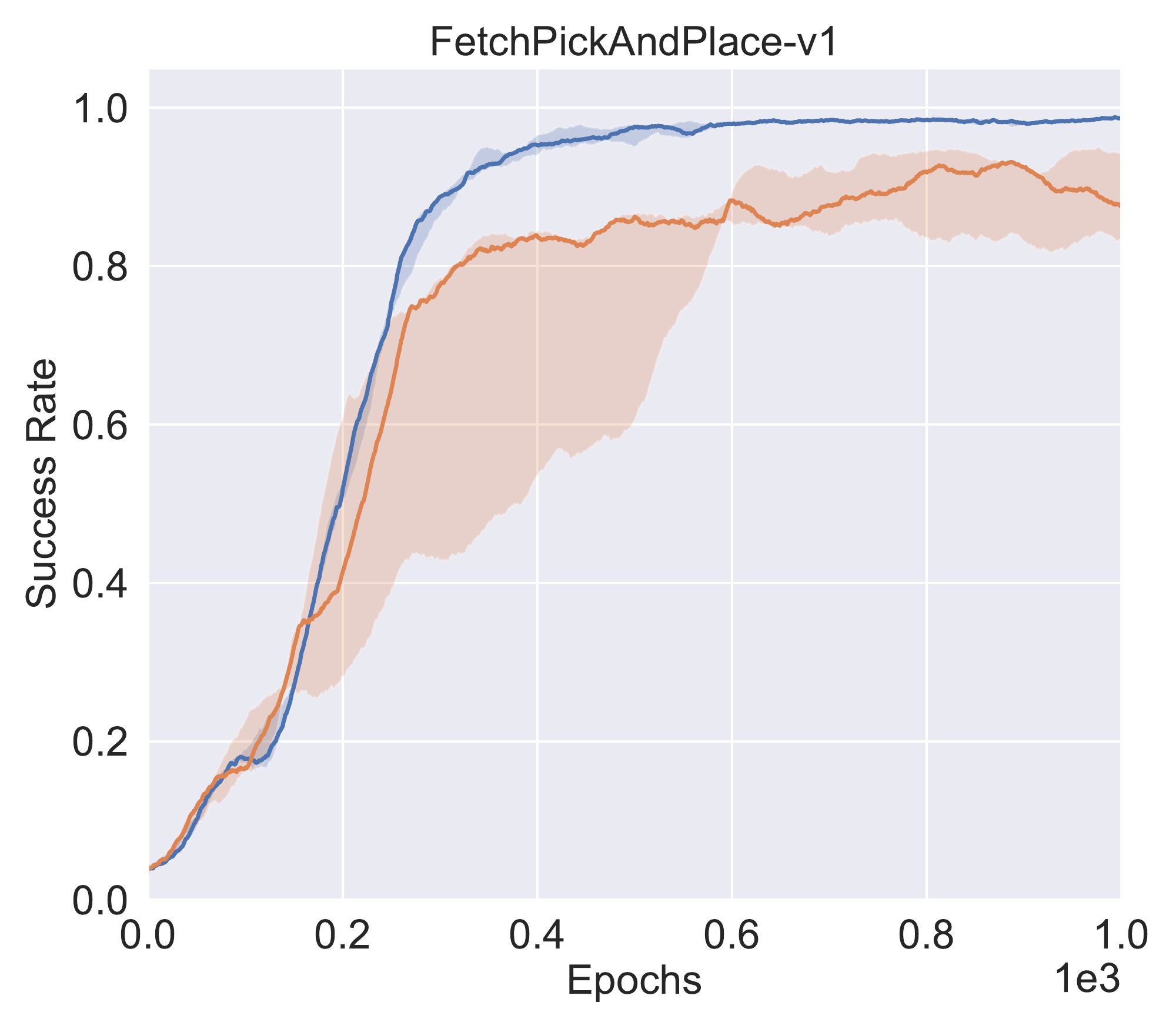}
  ({c}) FetchPickAndPlace
\endminipage
\minipage{0.5\textwidth}%
  \centering
  \includegraphics[width=\linewidth]{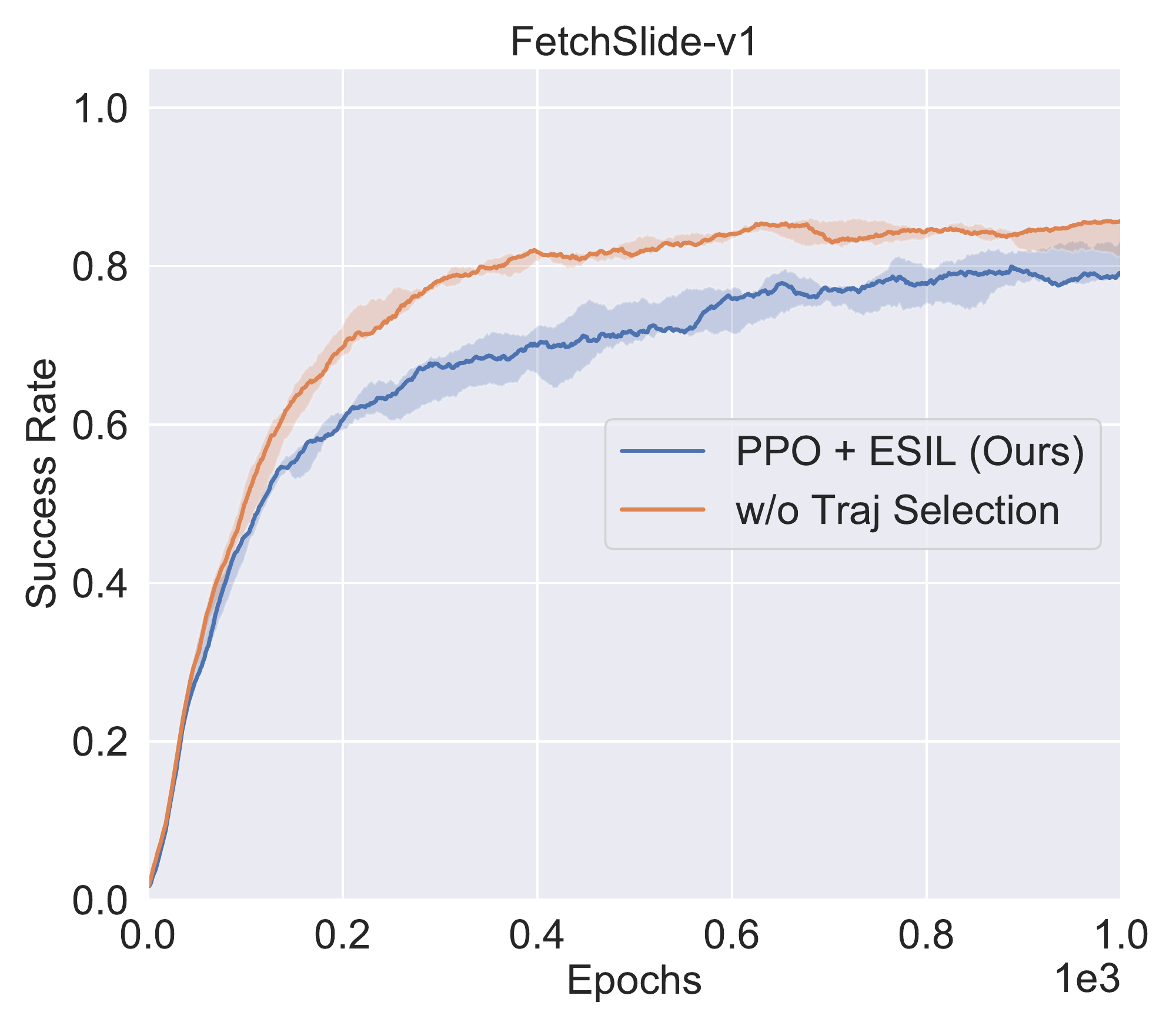}
  ({d}) FetchSlide
\endminipage\hfill
\caption{Results of ablation studies with or without using trajectory selection module on all Fetch environments.}
\label{fig:hs_compare}
\end{figure}

\subsubsection{Comparison to On-Policy Baselines}
{Figure~\ref{fig:baseline_compare}, PPO+ESIL} achieves reasonable results on all Fetch environments. {In contrast}, PPO, PPO+SIL and PPO+SIL+HER do not work on all tasks, with the exception of FetchReach. In comparison with the other selected tasks from the Fetch environments, FetchReach is relatively simple, because there is no object to be manipulated. For other tasks, it is quite difficult for the agent to achieve sufficient positive rewards during exploration, because of their rare occurrence. Although PPO+SIL utilises past good experiences to help exploration, it is still faced with the difficulty that past experiences do not easily achieve positive rewards. From the experiments ({see Figure~\ref{fig:baseline_compare}}), PPO+SIL (no hindsight) converges much more slowly than using PPO only. Attempting to use only the original trajectories for self-imitation learning leads to unsatisfactory performance. For PPO+SIL+HER (with no episodic update), the sampled transitions are modified into hindsight experiences, achieving better performance in the FetchReach and FetchSlide tasks. However, this transition-based method still cannot solve the other two manipulation tasks. In contrast, the proposed PPO+ESIL, through utilizing episodic hindsight experiences from failed trajectories, can achieve positive rewards quickly at the start of~training.

\subsubsection{Ablation Study of Trajectory Selection Module}
\begin{figure}[t!]
\centering
\minipage{0.5\textwidth}
  \centering
  \includegraphics[width=\linewidth]{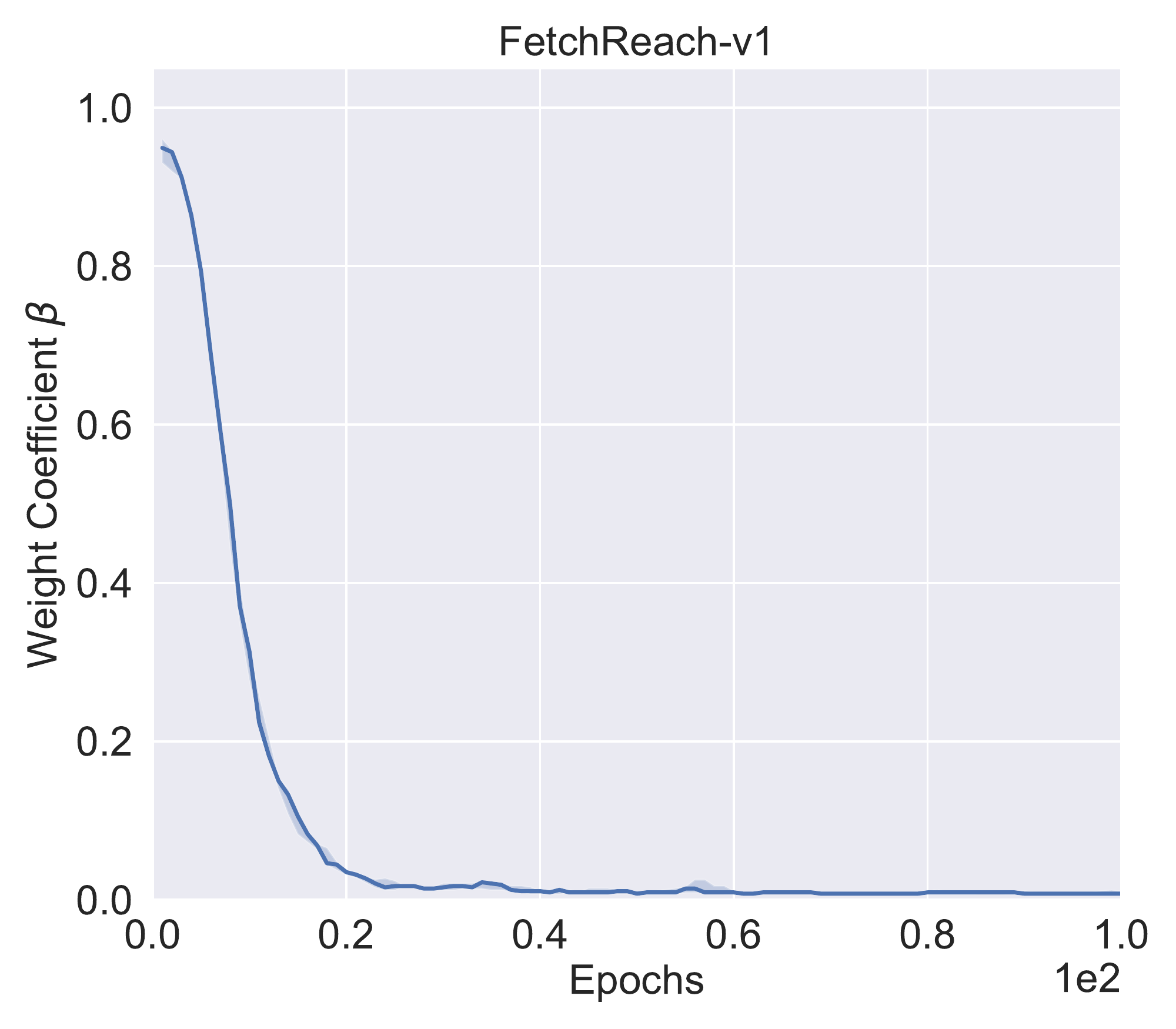}
  ({a}) FetchReach
\endminipage
\minipage{0.5\textwidth}%
  \centering
  \includegraphics[width=\linewidth]{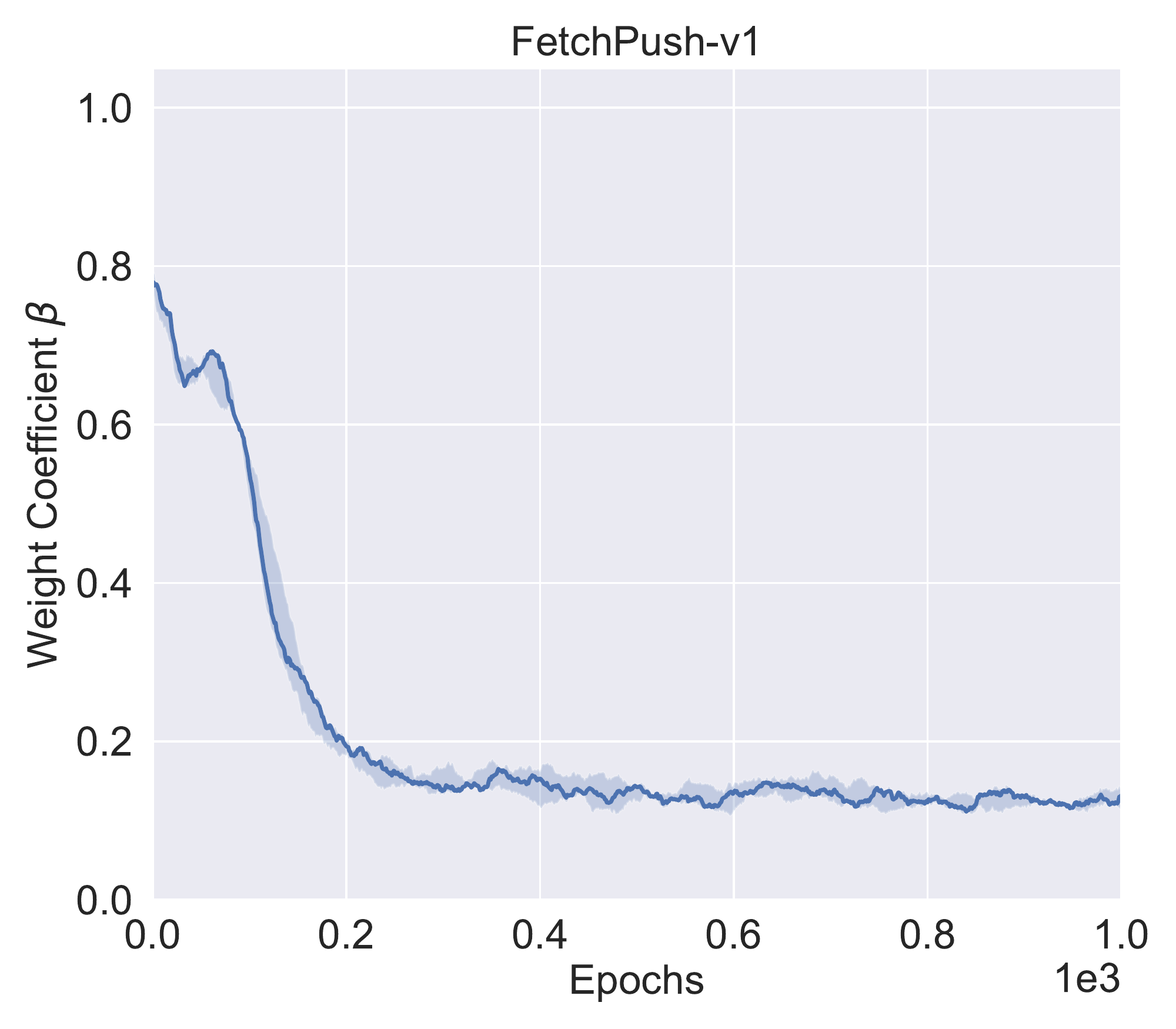}
  ({b}) FetchPush
\endminipage\hfill
\minipage{0.5\textwidth}%
  \centering
  \includegraphics[width=\linewidth]{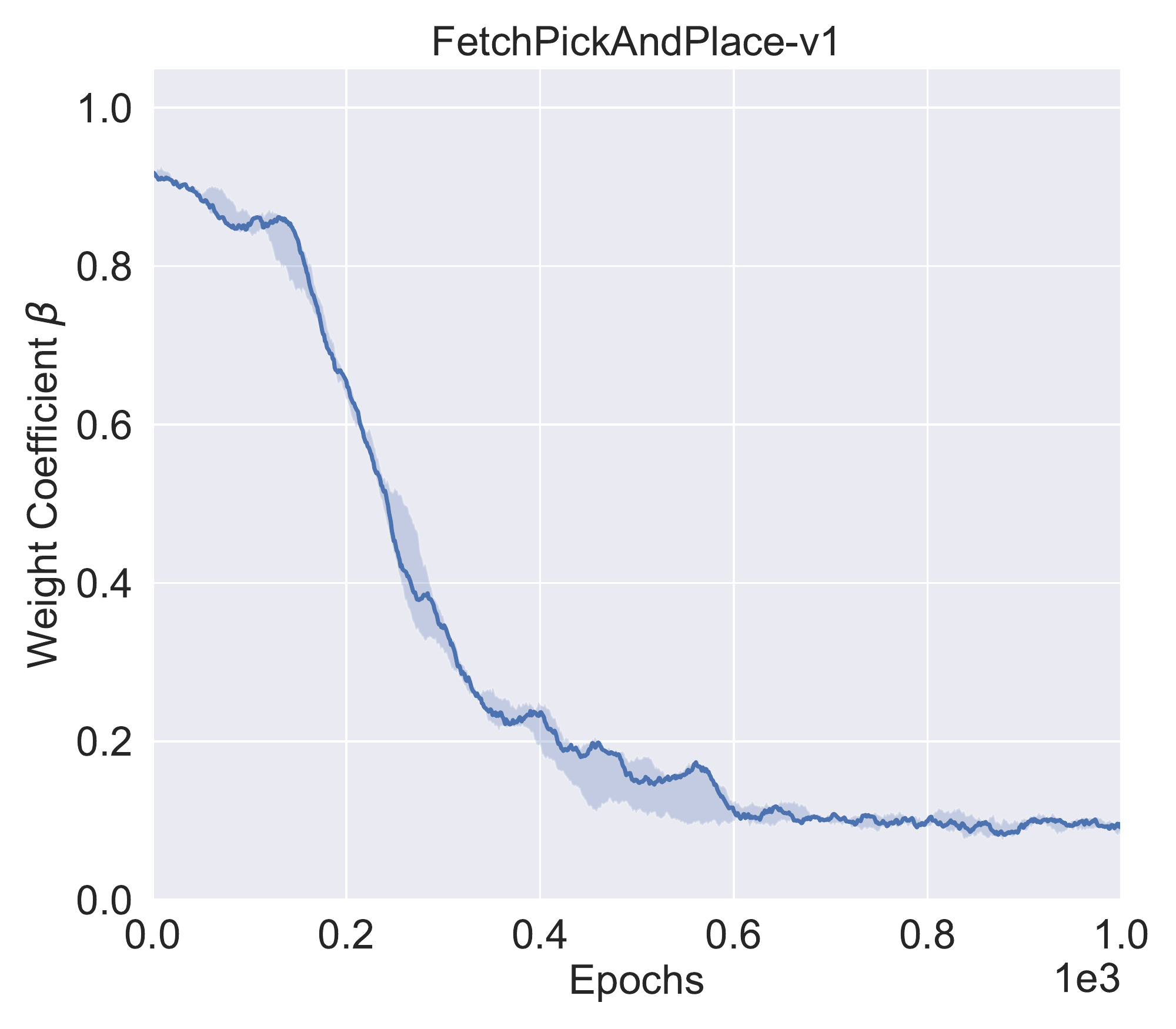}
  ({c}) FetchPickAndPlace
\endminipage
\minipage{0.5\textwidth}%
  \centering
  \includegraphics[width=\linewidth]{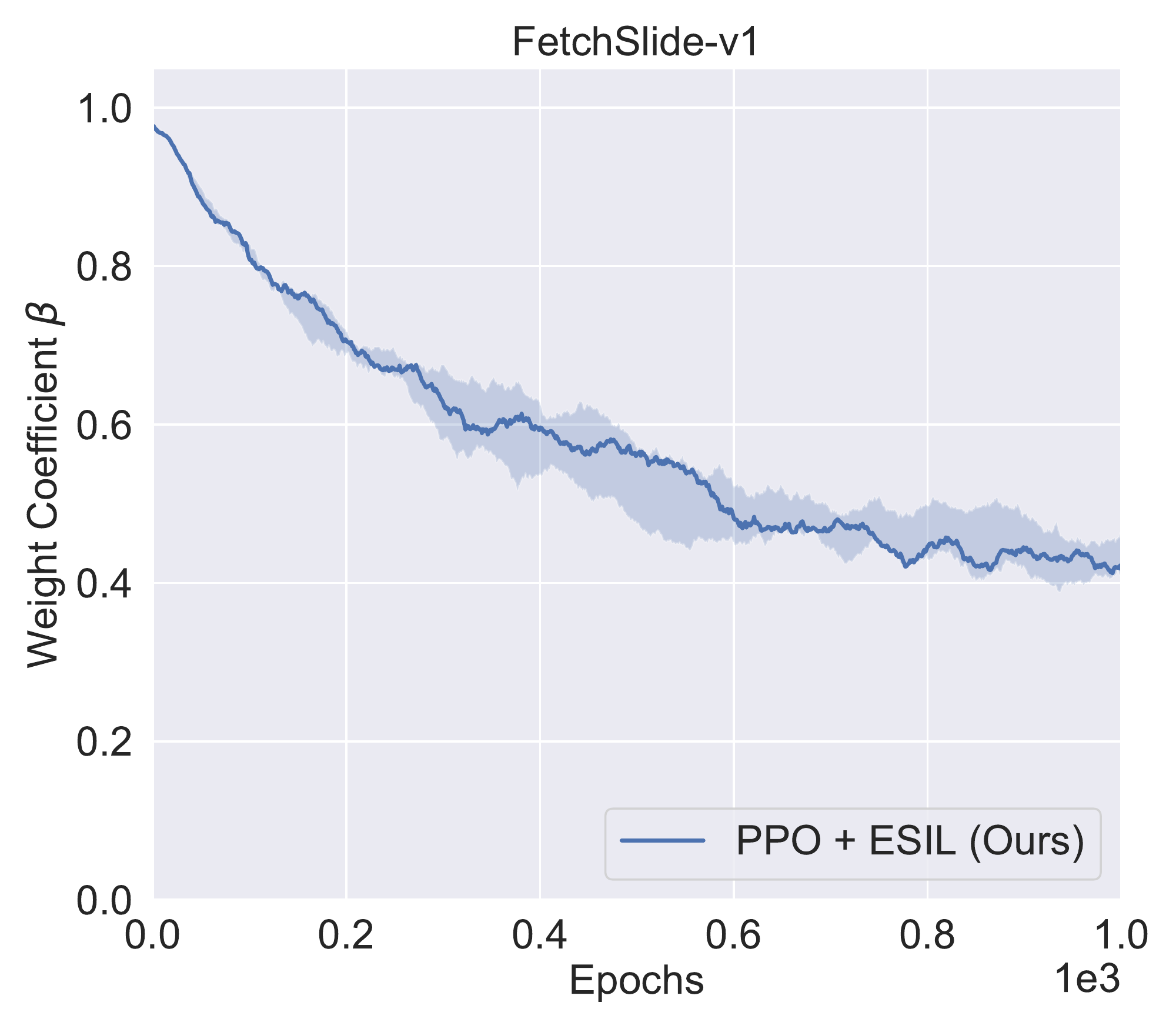}
  ({d}) FetchSlide
\endminipage\hfill
\caption{Variation in adaptive weight coefficient $\beta$ through the training on all Fetch environments.}
\label{fig:her_samples_compare}
\end{figure}

In order to investigate the effect of trajectory selection, {ablation studies are performed} to validate the selection strategy of our approach. {Figure~\ref{fig:hs_compare}}, when the trajectory selection module is not used, the~ performance of the agent increases at first, and then starts to decrease. This suggests that the agent starts to converge to a sub-optimal location. However, {Figure~\ref{fig:hs_compare}d}, for the FetchSlide task, the agent converges faster without the trajectory selection module, and has better performance. {This is likely to be because FetchSlide is the most difficult of the Fetch environments. During training, the agent is very unlikely to achieve positive rewards. Figure~\ref{fig:her_samples_compare} also indicates that the value of $\beta$ in FetchSlide is higher than values in other environments, which means the majority of hindsight experiences have higher returns than original experiences. Thus, using {\em more} hindsight experiences (without filtering) accelerates training at this stage.} Nonetheless, the trajectory selection module prevents the agent overfitting the hindsight experience in the other three tasks. Figure~\ref{fig:her_samples_compare}, shows the adaptive weight coefficient $\beta$ on all Fetch environments. When the trajectory selection module is used, the value of $\beta$ decreases with the increase in training epochs. This implies that the agent can achieve a greater proportion of the original goals in the latter stages of training, and fewer hindsight experiences are required for  self-imitation learning.
% It also shows that the performances of both agents still have tendencies to increase. So they still do not converge to their best performances. Thus it does not observe the agent without trajectory selection module has the performance drop in this task
\subsubsection{Comparison to Off-Policy Baselines}
\begin{figure}[t!]
\minipage{0.5\textwidth}
  \centering
  \includegraphics[width=\linewidth]{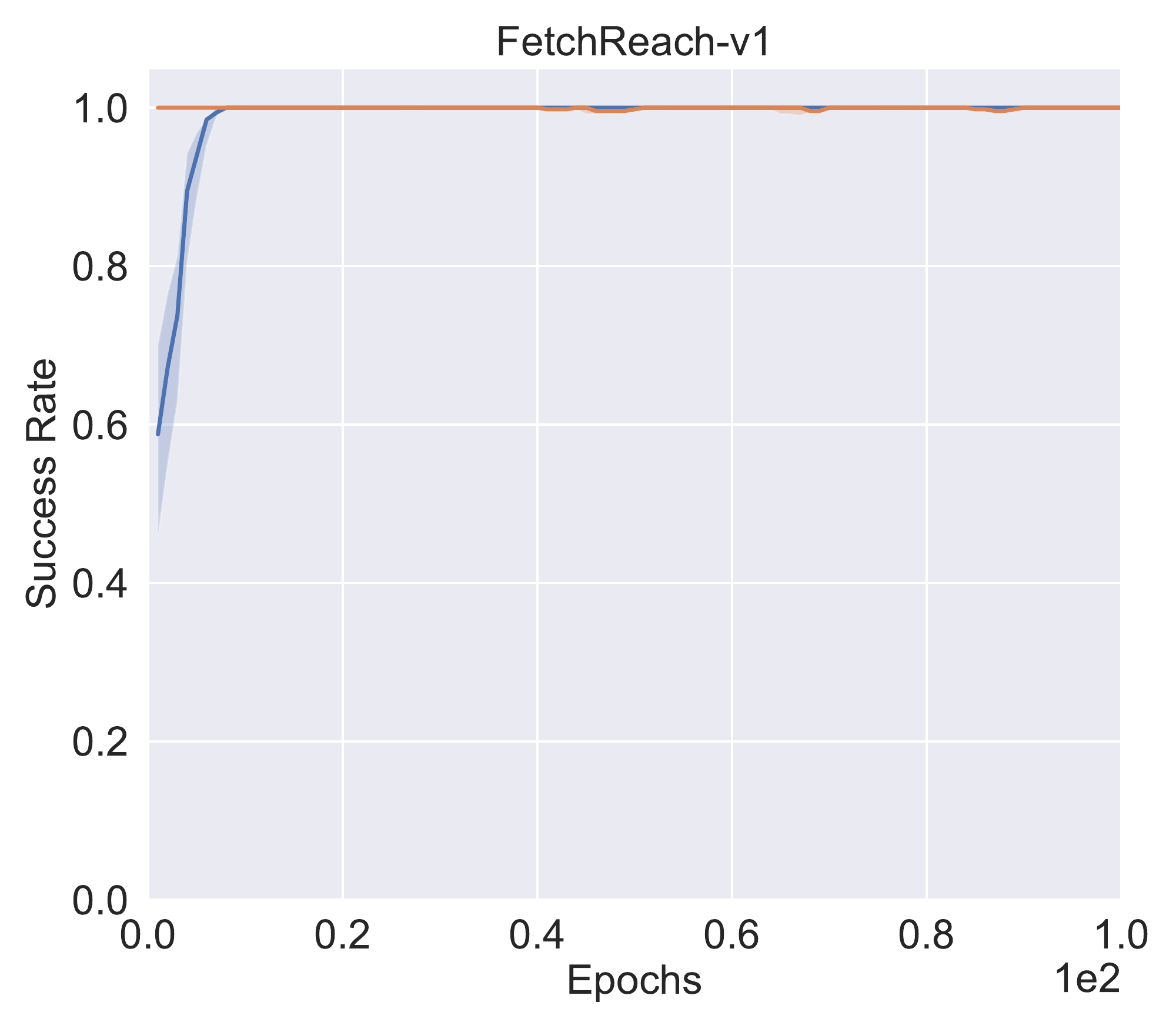}
  ({a}) FetchReach
\endminipage
\minipage{0.5\textwidth}%
  \centering
  \includegraphics[width=\linewidth]{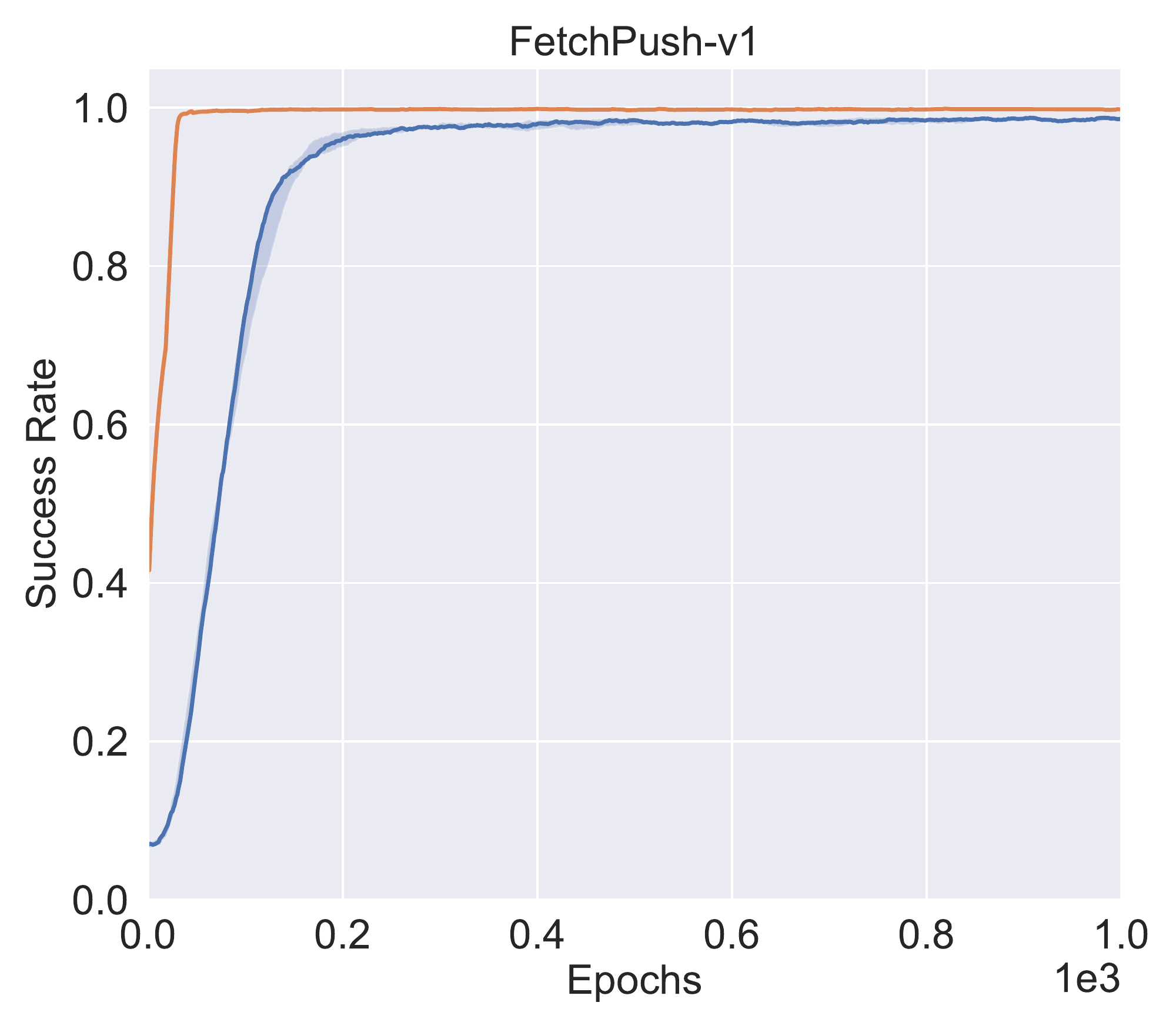}
  ({b}) FetchPush
\endminipage\hfill
\minipage{0.5\textwidth}%
  \centering
  \includegraphics[width=\linewidth]{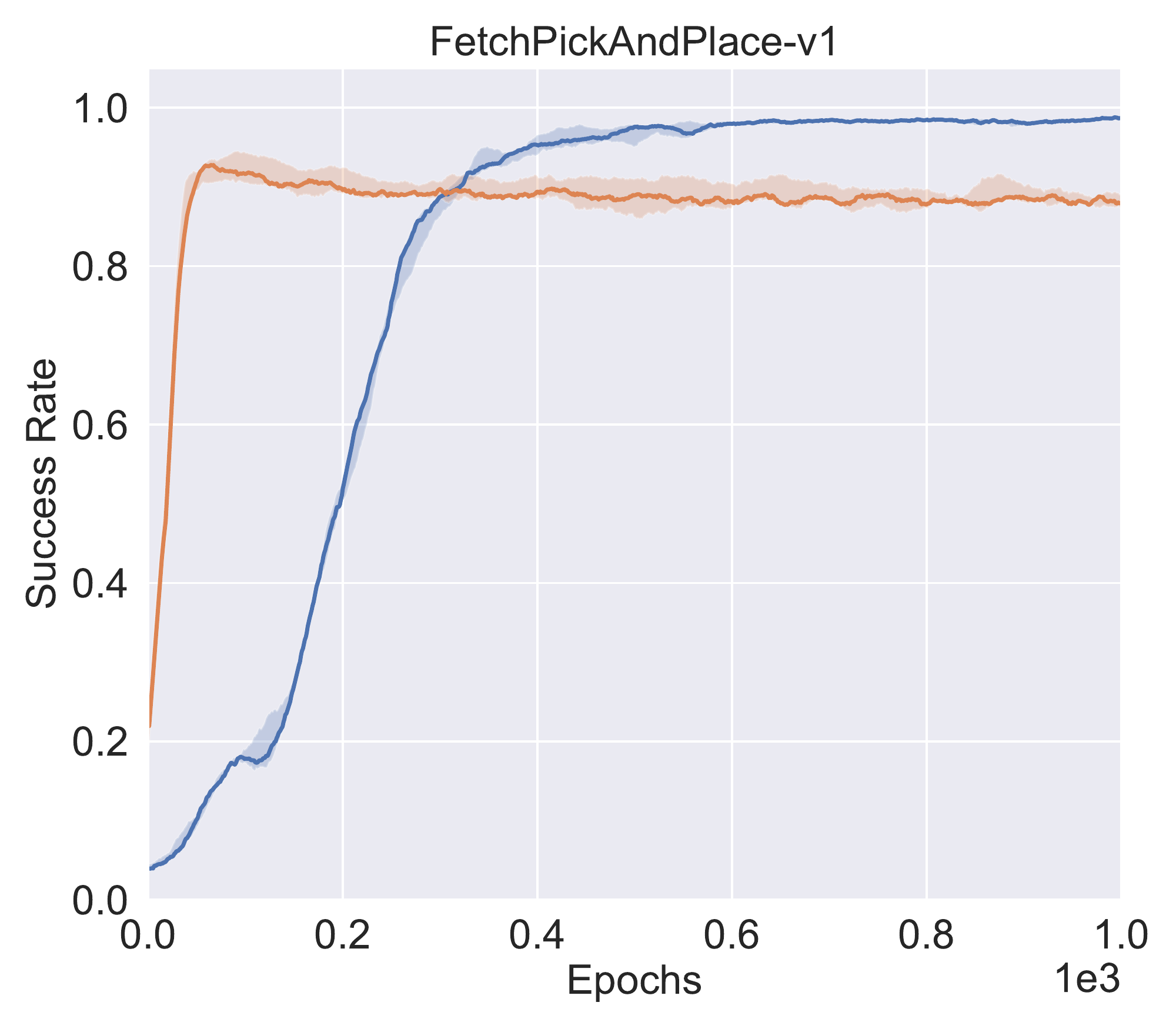}
  ({c}) FetchPickAndPlace
\endminipage
\minipage{0.5\textwidth}%
  \centering
  \includegraphics[width=\linewidth]{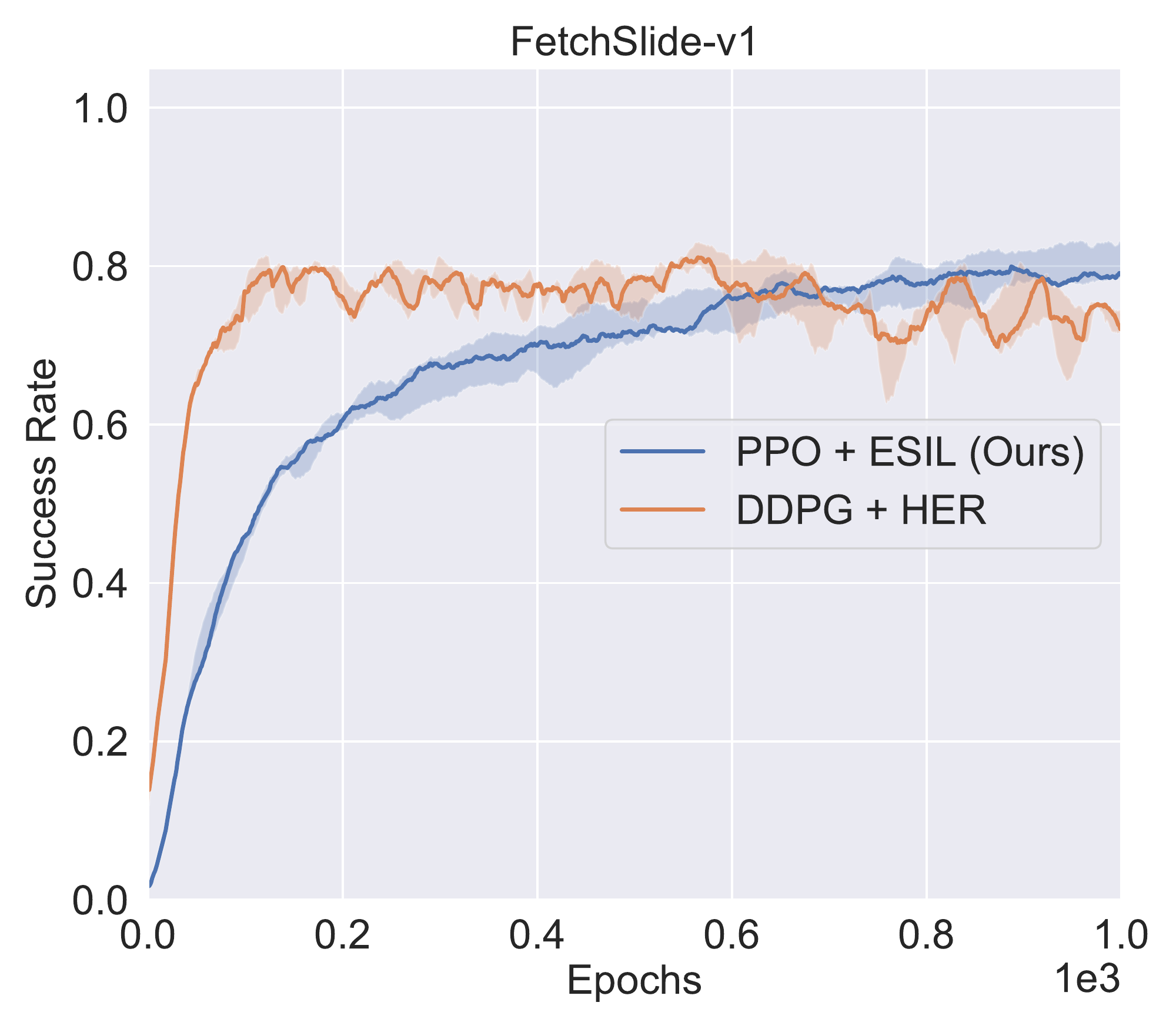}
  ({d}) FetchSlide
\endminipage\hfill
\caption{Results of comparison between PPO+ESIL and DDPG+HER on all Fetch environments.}
\label{fig:her_compare}
\end{figure}

Finally, {the proposed method is also compared} with a state-of-the-art off-policy algorithm: DDPG+HER. From {Figure~\ref{fig:her_compare}, it may be seen that} DDPG+HER converges faster than PPO+ESIL in all tasks. {However, PPO+ESIL obtains a similar performance to DDPG+HER. This is because DDPG+HER is an off-policy algorithm and uses a large number of hindsight experiences. A replay buffer is also employed to store  samples collected in the past. This approach has better sample efficiency than on-policy algorithms such as PPO.} Even so, {Figure~\ref{fig:her_compare}c} shows that PPO+ESIL still outperforms DDPG+HER in the FetchPickAndPlace task and the success rate is close to 1. {This suggests that PPO+ESIL approximates the characteristics of on-policy algorithms, which have low sample efficiency, but are able to obtain a comparable performance to off-policy algorithms in continuous control tasks~\cite{schulman2017proximal}.}

\subsection{Overall Performance}
{Table~\ref{tb:last}} shows the average success rate of the last 10 epochs during training of baseline methods and PPO+ESIL. The proposed ESIL achieves the best performance in four out of five tasks. However~PPO and PPO+SIL  only obtain reasonable results for the Empty Room and FetchReach tasks. With the assistance of HER, PPO+SIL+HER obtains a better performance in the FetchSlide task. {For the off-policy methods of DDPG+HER, all five tasks are achieved, but a better performance is  obtained than PPO+ESIL only in the FetchPush task}.

\begin{table}[t]
  \caption{Average success rate $\pm$ standard error in the last 10 epochs over five random seeds on all~environments {(\textbf{bold} indicates the best result among all methods).}}
  \label{tb:last}
  \centering
  \resizebox{\textwidth}{!}{
  \begin{tabular}{llllll}
    \toprule
    %\multicolumn{2}{c}{}                   \\
    %\cmidrule(r){1-2}
         & {Empty Room} & {Reach} & {Push} & {Pick} & {Slide}  \\
    \midrule
    PPO & 1.000 $\pm$ 0.000 & 1.000 $\pm$ 0.000& 0.070 $\pm$ 0.001 & 0.033 $\pm$ 0.001 & 0.077 $\pm$ 0.001 \\
    PPO + SIL & 0.998 $\pm$ 0.002 & 0.225 $\pm$ 0.016 & 0.071 $\pm$ 0.001 & 0.036 $\pm$ 0.002 & 0.011 $\pm$ 0.001 \\
    PPO + SIL + HER &0.996 $\pm$ 0.013 & 1.000 $\pm$ 0.000 & 0.066 $\pm$ 0.011 & 0.035 $\pm$ 0.004 & 0.276 $\pm$ 0.011 \\
    DQN + HER &1.000 $\pm$ 0.000 & - & - & - & - \\
    DDPG + HER & - & 1.000 $\pm$ 0.000 & \textbf{{0.996} %Please confirm whether this bold format is necessary? If not, please remove it. If it is necessary, please add the explanation.
 $\pm$ 0.001 %.
} & 0.888 $\pm$ 0.008 & 0.733 $\pm$ 0.013\\
    HPG &0.964 $\pm$ 0.012 & - & - & - & - \\
    \midrule
    PPO + ESIL (Ours) & \textbf{1.000 $\pm$ 0.000} & \textbf{1.000 $\pm$ 0.000}& 0.984 $\pm$ 0.003& \textbf{0.986 $\pm$ 0.002}& \textbf{0.812 $\pm$ 0.015}\\
    \bottomrule
  \end{tabular}}
\end{table}

\section{Conclusions}
\label{sec:conclusion}
This paper proposed a novel method for self-imitation learning (SIL), in which an on-policy RL algorithm uses episodic modified past trajectories, i.e., hindsight experiences, to update policies. Compared with standard self-imitation learning, episodic self-imitation learning (ESIL) has a better performance in continuous control tasks where rewards are sparse. As far as we know, it is also the first time that hindsight experiences have been combined with state-of-the-art on-policy RL algorithms, such as PPO, to solve relatively hard exploration environments in continuous action spaces.

{The experiments that we have conducted suggest that simply using self-imitation learning with the PPO algorithm, even with hindsight experience, leads to disappointing performance in continuous control Fetch tasks. In contrast, the episodic approach we take with ESIL is able to learn in these sparse reward settings. The auxiliary trajectory selection module and the adaptive weight $\beta$ help the training process to remove undesired experiences and balance the contributions to learning between the PPO term and the ESIL term automatically, and also increase the stability of training.} 

{Our experiments suggest that the selection module is useful to prevent overfitting to sub-optimal hindsight experiences, but also that it does not always lead to learning a better policy faster. Despite~this, selection filtering appears to support learning a useful policy in challenging environments. The experiments we have conducted to date have utilised relatively small networks, and it would be appropriate to extend the experiments to consider more complex observation spaces, and to actor/critic networks, which are consequently more elaborate.}

Future work includes extending {the proposed} method to support hierarchical reinforcement learning (HRL) algorithms for more complex manipulation control tasks, such as in-hand manipulation. Episodic self-imitation learning (ESIL) can also be applied to simultaneously learn sub-goal policies.

\begin{ack}
This work was partly supported by the Engineering and Physical Sciences Research Council [grant number: EP/J021199/1].
\end{ack}

\small
\bibliography{ref}
\bibliographystyle{unsrt}

\end{document}